\documentclass[review,number,12pt]{elsarticle}
\usepackage{latexsym}
\usepackage{amsmath}
\usepackage{amsfonts}
\usepackage{graphicx}
\usepackage{color}
\usepackage[normalem]{ulem}
\usepackage{soul}
\usepackage{setspace}
\usepackage{multirow}
\usepackage{subfigure}
\usepackage{bm}
\usepackage{appendix}
\usepackage{lineno}
\usepackage{subfig}
\usepackage{adjustbox}
\usepackage{booktabs}
\usepackage{array}
\usepackage{url}

\newcommand*\patchAmsMathEnvironmentForLineno[1]{
  \expandafter\let\csname old#1\expandafter\endcsname\csname #1\endcsname
  \expandafter\let\csname oldend#1\expandafter\endcsname\csname end#1\endcsname
  \renewenvironment{#1}
     {\linenomath\csname old#1\endcsname}
     {\csname oldend#1\endcsname\endlinenomath}}
\newcommand*\patchBothAmsMathEnvironmentsForLineno[1]{
  \patchAmsMathEnvironmentForLineno{#1}
  \patchAmsMathEnvironmentForLineno{#1*}}
\AtBeginDocument{
\patchBothAmsMathEnvironmentsForLineno{equation}
\patchBothAmsMathEnvironmentsForLineno{align}
\patchBothAmsMathEnvironmentsForLineno{flalign}
\patchBothAmsMathEnvironmentsForLineno{alignat}
\patchBothAmsMathEnvironmentsForLineno{gather}
\patchBothAmsMathEnvironmentsForLineno{multline}

\DeclareMathOperator\erf{erf}

\DeclareMathOperator{\vect}{vec}
\urlstyle{same}
}

\newcommand\xoutred{\bgroup \markoverwith{\textcolor{red}{\hbox to.35em{\hss/\hss}}}\ULon}

\ifpreprint
\fi

\journal{IEEE Access}
\begin{document}
\begin{frontmatter}

\title{Sobolev neural network with residual weighting as a surrogate in linear and non-linear mechanics}
\address[TUD]{Chair for Reliability Engineering, Fakultät Maschinenbau, TU Dortmund University, Leonhard-Euler-Strasse 5, 44227 Dortmund, Germany}
\address[TUDBM]{Lehrstuhl Baumechanik, Fakultät Architektur und Bauingenieurwesen, TU Dortmund University, August-Schmidt-Str. 8, 44227 Dortmund, Deutschland}
\author[TUD]{A.O.M. Kilicsoy \corref{corr_author}}
\author[TUDBM]{J. Liedmann}
\author[TUD]{M.A. Valdebenito}
\author[TUDBM]{F.-J. Barthold}
\author[TUD]{M.G.R. Faes}
\cortext[corr_author]{E-mail: ali.kilicsoy@tu-dortmund.de}

\begin{abstract}
Areas of computational mechanics such as uncertainty quantification and optimization usually involve repeated evaluation of numerical models that represent the behavior of engineering systems.
In the case of complex nonlinear systems however, these models tend to be expensive to evaluate, making surrogate models quite valuable.
Artificial neural networks approximate systems very well by taking advantage of the inherent information of its given training data.
In this context, this paper investigates the improvement of the training process by including sensitivity information, which are partial derivatives w.r.t. inputs, as outlined by Sobolev training.
In computational mechanics, sensitivities can be applied to neural networks by expanding the training loss function with additional loss terms, thereby improving training convergence resulting in lower generalisation error.
This improvement is shown in two examples of linear and non-linear material behavior.
More specifically, the Sobolev designed loss function is expanded with residual weights adjusting the effect of each loss on the training step.
Residual weighting is the given scaling to the different training data, which in this case are response and sensitivities.
These residual weights are optimized by an adaptive scheme, whereby varying objective functions are explored, with some showing improvements in accuracy and precision of the general training convergence.
\end{abstract}
\end{frontmatter}

\textit{Highlights:}
\begin{itemize}
\item Application of Sobolev trained neural network on linear and nonlinear mechanics
\item Evaluation of neural network as a surrogate for finite element simulation
\item Exploration of adaptive scheme of weighting methods for improved neural network training
\end{itemize}
\newpage

\section{Introduction}
\label{sec:introduction}
In multiple disciplines of engineering, various system analysis methods exist to predict their functional behavior.
Especially numerical techniques, such as finite element models (FEM), are a fundamental tool for various engineering tasks.
In particular, when quantifying uncertainty and performing optimization, repeated evaluations of these models are necessary~\cite{schueller2007treatment}.
Such repeated evaluations, however, scale poorly with system size and complexity.
Artificial neural networks (ANN) present an opportunity to replace time-intensive methods as a surrogate model~\cite{papadrakakis1996structural, hurtado2001neural} and have been successfully applied to the field of uncertainty quantification \cite{kantarakias2023sensitivity, bonnetquantifying, feng2022gradient}.
While the first version of ANNs, the Perceptron, dates far back~\cite{rosenblatt1958perceptron}, a surge of research has allowed ANNs to perform incredibly well on tasks previously too complex to solve by clear and definite methods with satisfactory results \cite{krizhevsky2012imagenet, devlin2019bert, Goodfellow_2014, lecun2015deep}.
In its most basic form, ANNs are capable of classification and regression, which in the case of the latter makes them general approximators \cite{hornik1989multilayer}.
There has also been research on intertwined processing between first principle models and neural networks \cite{mitusch2021hybrid, le2023finite}.
In the recent paper \cite{meethal2023finite}, the intrusive method of using the material and force tensor of a system used during FEM approximations, and applying these in the loss of a neural network is shown to evaluate the system better.
In any case, training a neural network can be a challenging process which has been approached in regards to the training algorithm \cite{dean2012large, kingma2014adam, berrada2020training}, training data \cite{wang2015learning, chen2014big} and loss function design \cite{raissi2017physics, raissi2019physics}.
In this work, specifically the adjustment of the loss function in order to introduce sensitivities, is of interest, as sensitivities bring additional information with them, which can be used to improve the training of models \cite{laurent2019overview, bhaduri2020usefulness}.
The use of gradient data to improve the performance of neural networks has been applied in a handful of papers and in the case of available sensitivity data, gradient-enhancement is an essential and simple step.
Usually, these sensitivities can be computed efficiently through adjoint methods, but these can be evaluated directly as well \cite{larson2013finite}.
The paper coining the term Sobolev training \cite{czarnecki2017sobolev}, shows the use of gradient data linked to Sobolev spaces, naming the models as Sobolev artificial neural network, or SANN.
The paper goes to prove the usability of gradient information through Sobolev space for a small, simple neural network.
Similar in essence, application of Lipschitz continuity during training can lead to an increase in robustness of the model, coined as Sobolev regularization \cite{gouk2020regularisation, finlay2019lipschitz}.
In another paper \cite{Yu_2022b}, the use of gradient data of a partial differential equation (PDE) when considering a physics informed neural network \cite{lu2021deepxde}, named gradient-enhanced PINN, or gPINN, improves the accuracy and training performance of a PINN, applying it on a few mathematical function examples.
In any case, Sobolev training is shown to produce better accuracy and training convergence than basic training, where only the model response is considered.
In regards to correctly weighting the Sobolev training losses, another paper expands on SANN by introducing weighting of the individual losses introduced by Sobolev training, by applying a fixed linear increase of the weight attributed to sensitivities, thereby improving performance, consequently naming it mSANN, or modified Sobolev artificial neural network \cite{bouhlel2020scalable}.
These expansions of ANNs allow for greater accuracy and faster training, taking ANNs a step further as surrogate models replacing established numerical methods \cite{grossmann2023can}.
In this work, the established Sobolev training is applied to approximate a mechanical problem of linear and non-linear nature, whereby a finite element model employing variational sensitivity analysis is used to compute responses and sensitivities needed for training of the ANN \cite{barthold2016, liedmann2020}.
More specifically, this gradient-enhanced neural network focuses on surrogate modelling the finite element model which uses variational sensitivity analysis of a mechanical system of linear and non-linear elasticity material behavior \cite{mythesis}.
Because various model designs are generally task specific, application to real physical tasks is important for further insights.
Additionally, application of Sobolev training with larger neural networks compared to e.g. the Sobolev training paper, but still small compared to deep neural networks, will provide new insights.
In contrast to PINNs, we only base the training of the Sobolev neural network on responses and their sensitivities that are obtained by performing variational finite element simulations, which is non-intrusive to the general neural network architecture.
For this, the loss function of the neural network is expanded with additional losses for each sensitivity, by which the model parameters are optimized in order for the backpropagated gradients to approximate the given sensitivities.
Lastly, in this work, this newly designed loss function consisting of multiple losses, will be improved by residual weights.
The residual weights will scale the individual loss terms which correspond to the response and sensitivities of the training data.
The goal of this residual weighting is the optimization of the Sobolev training convergence leading to faster training or lower approximation error.
This weighting task has been explored within mSANN, but also in other papers, where different algorithms and methods of optimization are applied \cite{mcclenny2020self, liu2021dual}.
In the same non-intrusive way as mSANN, the loss function will be expanded through residual weights corresponding to each individual loss.
However in this work, an adaptive scheme to optimize the residual weights is applied, which could potentially improve Sobolev training further, in contrast to general parameter tuning such as in mSANN.
This adaptive scheme optimizes the residual weights in a parallel process during training of the neural network, for their own defined objective function.
A set of objective functions are chosen based on prior beliefs of the system setup by the multiple individual losses.
The different methods are tested and then compared and analysed.
With this framework set, the paper will be structured as entailed in the following.
Section \ref{sec:Theory} will give a brief theoretical overview of the fundamentals this work is based on, such as the FEM and ANNs.
Section \ref{sec:Developments} of this work gives an overview of Sobolev training as it has already been done and then links it to the developments of residual weighting in detail.
Section \ref{sec:ToolsandApplication} describes the mechanical use case simulated by the FEM and the details on the programming of the neural networks.
Section \ref{sec:Results} presents the results of the residual weighting methods for further analysis.
The following section \ref{sec:AnalysisandDiscussion} covers the analysis and discussion of the results.
Section \ref{sec:Conclusion} concludes the work and its results, giving a short outlook on potential areas of further study.

\section{Theory}
\label{sec:Theory}

\subsection{General Remarks}
\label{sec:GeneralRemarks1}
This section will setup a foundation of the basic applied FEM and neural network.
This is done by elaborating the core formulation behind the variational finite element simulation providing the training data and the basic neural network calculations involved during training and prediction.
In regards to the neural network core terms are explained.

\subsection{Analysis Model}
\label{sec:AnalysisModel}
We make use of a finite element model which models a physical problem of linear or nonlinear material behavior.
The finite element method (FEM) is a numerical tool for solving underlying partial differential equations of engineering tasks.
In this work, the solution to the weak equilibrium equation for nonlinear elasticity in the reference configuration $\mathcal{B}_0$ is demanded 
\begin{align}\label{eq:weak_equilibrium}
    \begin{split}
	R(\bm{u},\bm{v}) &= \int_{\mathcal{B}_0} \bm{P}^{K}(\bm{u}) : \nabla\bm{v}\,dV \\ 
    &\quad - \int_{\mathcal{B}_0}\bm{b}_0 \cdot \bm{v}\,dV  \\ 
    &\quad - \int_{\mathcal{B}_0}\bm{t}_0 \cdot \bm{v}\,dA = 0,
    \end{split}
\end{align}
where $\bm{P}^{K}$ is the first Piola-Kirchhoff stress tensor, $\bm{b}_0$ and $\bm{t}_0$ denote body and traction forces, respectively.
Further, $\bm{u}$ denotes the vector of primary displacements and $\bm{v}$ is the vector of test functions.
Due to the nonlinear system response, a Newton-Raphson procedure is employed to iteratively solve Eq.~(\ref{eq:weak_equilibrium}), which implies its linearization
\begin{equation}\label{eq:weak_form_linearization}
	R(\bm{u},\bm{v}) + \delta_u R(\bm{u,\bm{v}; \Delta\bm{u}}) = 0.
\end{equation}
In the FEM a system is approximated through various techniques.
In a process called discretization, the observed system geometry is subdivided into multiple simple sections called finite elements, which make up the complete system mesh.
This allows to numerically solve the weak form of equilibrium.
The finite element system is set up through the use of known shape functions - traditionally chosen as polynomials - for the approximation of geometry, displacements and test functions.
Following the isoparametric concept, the same shape functions are used for all approximations in each finite element.
With the discrete approximations
\begin{equation}
	\begin{aligned}
		R(\bm{u},\bm{v}) \approx \boldsymbol{v}^{T}\boldsymbol{R} && \textbf{and } && \delta_u R(\bm{u},\bm{v},\Delta\boldsymbol{u}) \approx \boldsymbol{v}^{T} \boldsymbol{K} \Delta\boldsymbol{u},
	\end{aligned}
\end{equation}
where $\boldsymbol{R}$ and $\boldsymbol{K}$ 
denote the discrete residual vector and tangent stiffness matrix, respectively, and excluding the trivial solution $\boldsymbol{v} = \boldsymbol{\mathsf{0}}$, the discrete version of Eq.~(\ref{eq:weak_form_linearization}) solved for the unknown displacement increment vector $\Delta\boldsymbol{u}$ in each Newton-Raphson iteration reads
\begin{equation}
	\boldsymbol{K} \Delta\boldsymbol{u} = -\boldsymbol{R}.
\end{equation}%
Further details on finite element analysis (FEA) can be found e.g. in \cite{Bathe1990,Wriggers2001,Zienkiewicz2000a,Zienkiewicz2005finite,Hughes2012} to name a few.
In design sensitivity analysis, generally, the change of a response function $f(\boldsymbol{u}(\boldsymbol{x}),\boldsymbol{x})$ w.r.t. a change in chosen design or model parameters $\boldsymbol{x}$ defining the model design shall be quantified.
Utilising variational sensitivity analysis, cf. e.g. \cite{barthold2016, liedmann2020}, this change can be written in variational form
\begin{equation}\label{eq:arb_response_function}
	\delta f = \delta_u f + \delta_x f = \left[\frac{\partial f}{\partial \boldsymbol{u}}\right] \delta\boldsymbol{u} + \left[\frac{\partial f}{\partial \boldsymbol{x}}\right]\delta\bm{x}.
\end{equation}%
Further, considering that a design change $\delta \bm{x}$ must not violate the equilibrium of the system, i.e.  
\begin{equation*}
	\delta R(\bm{u},\bm{v},\delta\bm{u},\delta\bm{x}) = \delta_u R(\bm{u},\bm{v},\delta\bm{u}) + \delta_x R(\bm{u},\bm{v},\delta\bm{x}) = 0,
\end{equation*}%
and approximate both variations using
\begin{equation}
	\begin{aligned}
	\delta_u R(\boldsymbol{u},\boldsymbol{v},\delta\boldsymbol{u}) \approx \boldsymbol{v}^{T} \boldsymbol{K}\delta \boldsymbol{u} && \text{and} && \delta_x R(\boldsymbol{u},\boldsymbol{v},\delta\boldsymbol{x}) \approx \boldsymbol{v}^{T} \boldsymbol{P}\delta \boldsymbol{x},
	\end{aligned}
\end{equation}
where $\boldsymbol{P}$ represents the tangent pseudoload matrix, the total response sensitivity can be identified by rearranging the discrete total variation of the weak equilibrium condition, viz.
\begin{equation}\label{eq:feasible_design_constr}
	\begin{aligned}
	\boldsymbol{K}\delta\boldsymbol{u} = -\boldsymbol{P}\delta\boldsymbol{x}.
 \end{aligned}
\end{equation}%
This procedure is commonly referred to as the direct differentiation method and results in the computation of the sensitivity matrix $\boldsymbol{S}$ by solving Eq.~(\ref{eq:feasible_design_constr}) for the unknown displacement variations, i.e 
\begin{equation}\label{eq:response_sensitivity}
\delta\boldsymbol{u} = -\boldsymbol{K}^{-1}\boldsymbol{P}\delta\boldsymbol{x} = \boldsymbol{S}\delta\boldsymbol{x}.
\end{equation}%
The response function of interest in this study is the so called stress triaxiality defined as the ratio of the mean stress $\sigma_{\mathsf{Mean}}$ and equivalent von Mises stress $\sigma_{\mathsf{Mises}}$
\begin{equation}
	T = \frac{\sigma_{\mathsf{Mean}}}{\sigma_{\mathsf{Mises}}} = \frac{\operatorname{tr}\boldsymbol{\sigma}}{3 \sqrt{3 J_2}}, 
\end{equation}%
where $J_2$ is the second deviatoric stress invariant and $\operatorname{tr}\boldsymbol{\sigma}$ is the stress tensor.
Therefore, it characterizes the current stress state as follows.
A high positive stress triaxiality corresponds to a tensile stress state, while negative values indicate a compression stress state.
A low magnitude corresponds to a shear dominated stress state. \\
Utilizing the above described variational method, the discrete sensitivity relation of the stress triaxiality can be derived to
\begin{equation}
	\delta T = \left[\frac{\partial T}{\partial \sigma_{\mathsf{Mean}}} \frac{\partial \sigma_{\mathsf{Mean}}}{\partial \boldsymbol{\sigma}} + \frac{\partial T}{\partial \sigma_{\mathsf{Mises}}}\frac{\partial \sigma_{\mathsf{Mises}}}{\partial \boldsymbol{\sigma}} \right] \delta \boldsymbol{\sigma},
\end{equation}%
where the determination of the partial derivatives are straight forward and the total variation of the stress tensor can by means of Eq.~(\ref{eq:arb_response_function}) and Eq.~(\ref{eq:response_sensitivity}) be identified to
\begin{equation}
	\delta\boldsymbol{\sigma} = \left[\frac{\partial \boldsymbol{\sigma}}{\partial\boldsymbol{u}} \boldsymbol{S} + \frac{\partial \boldsymbol{\sigma}}{\partial \boldsymbol{x}} \right] \delta\boldsymbol{x}.
\end{equation}%
Further details on derivation and implementation can be found e.g. in the referenced literature.%
A nonlinear compressible Neo-Hookean strain energy function of the form
\begin{equation}
	W^{NH} = \frac{1}{2} \mu (I_c - 3 - 2 \log(J)) + \frac{1}{2} \lambda (J - 1)^2
\end{equation}
is used for the computation of the stress response, where $\mu$ and $\lambda$ are the elastic Lam\'e parameters, $I_c = \operatorname{tr}(\bm{C}) = \operatorname{tr}{(\bm{F}^{T}\bm{F})}$ is the first invariant of the left Cauchy-Green deformation tensor and $J = \det(\bm{F})$ denotes the determinant of the deformation gradient $\bm{F}$.

\paragraph{Remark} Note that within the work at hand the design parameters are chosen as geometric control point coordinates of a mesh controlling Bezier surface.
However, within the FEM the only geometric information available is the nodal mesh coordinates.
Therefore, a design velocity matrix is used to connect the nodal mesh coordinates with the underlying Bezier geometry description, cf. e.g. \cite{Barthold2008,Liedmann2017,Liedmann2018a}.

\subsection{Artificial Neural Networks}
\label{sec:ArtificialNeuralNetworks}
The neural network models in this paper have a basic structure and design of a regression feedforward model.
According to the Universal Approximation Theorem \cite{hornik1989multilayer} and its subsequent variations, neural networks are capable of approximating any continuous functions.
ANNs have been proven to provide accurate results empirically.
Following, the design parameters are referred to as the inputs $\bm{x}$ and the triaxiality $\bm{T}$ as true outputs $\bm{y}$.
\subsubsection{Preprocessing}
\label{sec:Preprocessing}
In supervised training, each input $\bm{x}_d$ is paired with a true response $\bm{y}_d$ for a given training data size $d$, whereby input and output usually contain multiple elements, denoted by subscript $e$. 
In general, the training data is split into multiple parts, which are called minibatches and each training iteration uses a single minibatch in order to minimize overfitting. The order of these minibatches is shuffled during training which adds a stochastic effect to the training process.
When applying training data to a neural network, it is generally recommended to preprocess the data to avoid general pitfalls, such as exploding and vanishing gradients.
In certain cases, trimming the data of outliers for numerical instability and normalization or standardization are important tools for training neural networks.
Input and response data $\bm{x}_d, \bm{y}_d$ can be standardized by using the mean values $\bm{\bar{x}}, \bm{\bar{y}}$ and dividing by the corresponding standard deviations of each element $\bm{\sigma_{\bm{x}}}, \bm{\sigma_{\bm{x}}}$ respectively per Hadamard division:
\begin{align}
    \bm{x}_{d,s} = \frac{\bm{x}_d - \bm{\bar{x}}}{\bm{\sigma_{\bm{x}}}}
\end{align}
\begin{align}
    \bm{y}_{d,s} = \frac{\bm{y}_d - \bm{\bar{y}}}{\bm{\sigma_{\bm{y}}}}
\end{align}
Generally, it is necessary to preprocess the data with the to be approximated system in mind.
The following sections will leave out subscript $d, e, s$  for the sake of simplicity.
\subsubsection{Forward Pass}
\label{sec:ForwardPass}
For all equations relating to neural networks in this section, the number of neurons is defined as $n$.
To provide a better overview in equations, the current layer from which anything is referenced is denoted as $l$, the following layer as $l+1$ and the previous layer is denoted as $l-1$.
In addition, layers are always denoted as superscripts in brackets, that is $(\cdot)^{[\cdot]}$.
The neurons are referenced with $i$, when referring to the current layer, and with $j$ when referring to the previous layer.
This numbering is denoted as subscripts, whereby in general $j$ follows after $i$ in the subscript. The linear term or the weighted sum, of a neuron, is called $z$, whereby $w$ denotes a weight and $b$ denotes a bias.
The nonlinear output of a node is called $o$, whereby the nonlinear function applied in it, is notated as $a(\cdot)$.
The input training data is referenced as $x$, the response training data or the true output as $y$ and the neural network model output as $\hat{y}$.
Every $x$ is paired with a $y$, as supervised training is applied to the neural network models.

However for the sake of overview, data size and minibatches will not be considered in the following equations and they apply to a single data pair of $x, y$.
Still, every equation is applicable to multiple training data samples for a single iteration by following the defined equations in this section for each training data pair of the minibatch and simply taking the average of the final result, the loss $\mathcal{L}$.
First, we define the weight matrices $\bm{W}^{[l]} \in R^{\, n^{[l]} \times n^{[l-1]}}$ for any layer $l$ as:
\begin{align}
    \bm{W}^{[l]} = \begin{bmatrix}
    w^{[l]}_{1,1} & \hdots &  w^{[l]}_{n^{[l-1]},1}\\
    \vdots & \ddots & \vdots \\
    w^{[l]}_{1,n^{[l]}} & \hdots & w^{[l]}_{n^{[l-1]},n^{[l]}}
    \end{bmatrix}
\end{align}
The bias vector $\bm{b}^{[l]} \in R^{\, n^{[l]} \times 1}$ follows the following pattern for any layer $l$:
\begin{align}
    \bm{b}^{[l]} = \begin{pmatrix}
    b^{[l]}_{1,1} \\
    \vdots \\
    b^{[l]}_{1,n^{[l]}}
    \end{pmatrix}
\end{align}
The output vectors $\bm{o}^{[l]} \in R^{\, n^{[l]}
}$ for any layer are defined as:
\begin{align}
    \bm{o}^{[l]} = \begin{pmatrix}
    o^{[l]}_{1}\\
    \vdots\\
    o^{[l]}_{n^{[l]}}
    \end{pmatrix}
\end{align}
The vectors $\bm{z}^{[l]} \in R^{\, n^{[l]}
}$ containing the linear terms are defined as:
\begin{align}
    \bm{z}^{[l]} = \bm{W}^{[l]} \cdot \bm{o}^{[l-1]} + \bm{b}^{[l]}
\end{align}
whereby the vectors $\bm{o}^{[l-1]}$ contain the nonlinear outputs of the neurons for the respective layer $l-1$.
$\bm{W} \in R^{\, n^{[l]} \times n^{[l-1]}}$ is the weight matrix, $\bm{b} \in R^{\, n^{[l]} \times 1}$ is the corresponding bias vector.
Written-out for each element, the equation changes to:
\begin{align}
    z^{[l]}_i = \sum_{j=1}^{n^{[l-1]}} o^{[l-1]}_j \cdot w^{[l]}_{ij} + b^{[l]}_{ij} \ ,\! \textrm{with }\ i = \{1,..,n^{[l]}\}
\end{align}
Each element of the output vector $\bm{o}^{[l]}$ is defined as a predefined activation function $a(\cdot)$ applied to the corresponding linear vectors $\bm{z}^{[l]}$, element-wise:
\begin{align}
    \bm{o}^{[l]} = a(\bm{z}^{[l]})
\end{align}
Written-out, the expression for each element is:
\begin{align}
    o^{[l]}_i = a(z^{[l]}_i),\! \textrm{~with} \ i = \{1,..,n^{[l]}\}
\end{align}
In the case of the input layer $l = 0$ and the output layer $l = L$ the output vector $\bm{o}$ is the input vector $\bm{x}$ or the output vector $\bm{\hat{y}}$, respectively:
\begin{align}
     \bm{x} = \begin{pmatrix}
        x_1 \\
        \vdots \\
        x_{n^{[0]}}
    \end{pmatrix}
    = \bm{o}^{[0]},\! \textrm{~with} \ l = 0
\end{align}
\begin{align}
     \bm{\hat{y}} = \begin{pmatrix}
        \hat{y}_1 \\
        \vdots \\
        \hat{y}_{n^{[L]}}
    \end{pmatrix}
    = \bm{o}^{[L]},\! \textrm{~with} \ l = L
\end{align}
During the training process a neural network compares its own produced response vector $\bm{\hat{y}}$ with the true response vector $\bm{y}$ with the help of a loss function $\mathcal{L}$.
Over the training process, the neural network model ideally converges this loss towards zero and its loss serves as a performance metric of the model's accuracy to approximate the training data.
The loss function $\mathcal{L}$ is defined as the sum of all losses, which is simply the response loss $\mathcal{E}_{\mathsf{Response}}$:
\begin{align}
    \mathcal{L} = \mathcal{E}_{\mathsf{Response}}
\end{align}
The response loss $\mathcal{E}_{\mathsf{Response}}$ is defined as the half-mean squared error of the model's response $\bm{\hat{y}}$ to the true response $\bm{y}$, whereby $\lVert \cdot \rVert$ denotes the Euclidean vector norm:
\begin{align}
    \mathcal{E}_{\mathsf{Response}} = \frac{1}{2} \cdot \lVert\bm{\hat{y}} - \bm{y}\rVert^2 = \sum_{i=1}^{n^{[L]}} \frac{1}{2} \cdot (\hat{y}_i - y_i)^2
\end{align}
\subsubsection{Backpropagation}
\label{sec:Backpropagation}
For the training process, it is necessary to optimise the model parameters, the weights and biases, referred to as $\theta$, consisting of the weights and biases of the neural network.
The neural network needs to identify the gradient of the loss with respect to each model parameter $\frac{\partial \mathcal{L}}{\partial \theta}$, for the minimization of the loss $\mathcal{L}$.
To compute all gradients for the model parameter update step, the backpropagation follows after the forward pass.
The forward pass populates each parameter of the neural network model and gives the current model response $\bm{\hat{y}}$.
$\theta$ means both weights $w$ or biases $b$ of its respective layer.
To compute the gradients with respect to the model parameters of layer $l = p$, automatic differentiation is used, which applies the chain rule during backpropagation to evaluate each partial derivative step by step.
The general formulation for a neural network with a single neuron in each of its layers defines the backpropagation for the partial derivatives $\frac{\partial \mathcal{L}}{\partial \bm{\theta}^{[p]}}$ of the loss $\mathcal{L}$ with respect to the model parameters $\bm{\theta}^{[p]}$ per chain rule as:
\begin{align}
    \begin{split}
        \frac{\partial \mathcal{L}}{\partial \theta^{[p]}} &= \frac{\partial \mathcal{L}}{\partial \hat{y}} \frac{\partial \hat{y}}{\partial o^{[L]}} \frac{\partial o^{[L]}}{\partial z^{[L]}} \frac{\partial z^{[L]}}{\partial o^{[L-1]}} \dots \frac{\partial o^{[p]}}{\partial z^{[p]}} \frac{\partial z^{[p]}}{\partial \theta^{[p]}}\ \\
        &\textrm{~with} \ \ 1 \leq p \leq L     
    \end{split}
\end{align}
Each partial derivative can then be evaluated following these fundamental derivatives.
For $\frac{\partial \mathcal{L}}{\partial \hat{y}}$ and $\frac{\partial \hat{y}}{\partial o^{[L]}}$:
\begin{align}
    \frac{\partial \mathcal{L}}{\partial \hat{y}} = \hat{y} - y \ \textrm{and} \ \frac{\partial \hat{y}}{\partial o^{[L]}}  = 1
\end{align}
And the repeated partial derivatives through the layers:
\begin{align}
    \frac{\partial o^{[l]}}{\partial z^{[l]}} = a'(z^{[l]}) \ \textrm{and} \ \frac{\partial z^{[l]}}{\partial o^{[l-1]}} = w^{[l]}
\end{align}
And for target layer $l=p$ we have:
\begin{align}
    \frac{\partial z^{[p]}}{\partial \theta^{[p]}} =
    \begin{cases} 
    o^{[p-1]} & \text{if  }  \theta^{[p]} = w^{[p]} \\
    1 & \text{if  } \theta^{[p]} = b^{[p]} 
    \end{cases}
\end{align}
Whereas in the last equation the distinction occurs on whether the model parameter $\theta$ is a weight $w$ or bias $b$.
When considering the order of operation in a feedforward neural network, each partial derivative with respect to model parameters of layer $p$ is preceded by the partial derivative with respect to model parameters of its following layer $p+1$.
Next, we define the partial derivative with matrices and vectors, since we are dealing with more than a single neuron in the layers.
With respect to layer $l$ we have the partial derivative with respect to the nonlinear output $\bm{\delta}^{[l]}$:
\begin{align}
     \bm{\delta}^{[l]} = \frac{\partial \bm{z}^{[l]}}{\partial \bm{o}^{[l-1]}} \otimes    \bm{\delta}^{[l+1]} \odot \frac{\partial \bm{o}^{[l]}}{\partial \bm{z}^{[l]}}
\end{align}
The operator $\odot$ denotes the Hadamard product, which is an element-wise multiplication, whereas the operator $\otimes$ denotes basic matrix multiplication. 
The terms are defined as follows:
\begin{align}
    \frac{\partial \bm{z}^{[l+1]}}{\partial \bm{o}^{[l]}} = (\bm{W}^{[l+1]})^T \in R^{n^{[l]} \times n^{[l+1]}}
\end{align}
\begin{align}
    \frac{\partial \bm{o}^{[l]}}{\partial \bm{z}^{[l]}} = a'(\bm{z}^{[l]}) \in R^{n^{[l]} \times 1}
\end{align}
The $\bm{\delta}^{[L]}$ at the last layer $L$ is defined as:
\begin{align}
     \bm{\delta}^{[L]} =  \frac{\partial \mathcal{L}}{\partial \bm{\hat{y}}} \odot \frac{\partial \bm{\hat{y}}}{\partial \bm{z}^{[L]}} \in R^{n^{[L]} \times 1}
\end{align}
From here on, we can define the partial derivatives with respect to model parameters in the following:
\begin{align}
     \frac{\partial \mathcal{L}}{\partial W^{[l]}} = \bm{\delta}^{[l]} \otimes \frac{\partial \bm{z}^{[l]}}{\partial \bm{W}^{[l]}} \in R^{n^{[l]} \times n^{[l-1]}}
\end{align}
In this term, we have defined $\frac{\partial \bm{z}^{[l-1]}}{\partial W^{[l]}}$ as:
\begin{align}
     \frac{\partial \bm{z}^{[l]}}{\partial \bm{W}^{[l]}} = (\bm{o}^{[l-1]})^T
\end{align}
For the biases, we have instead:
\begin{align}
     \frac{\partial \mathcal{L}}{\partial \bm{b}^{[l]}} = \bm{\delta}^{[l]} \in R^{n^{[l]} \times 1}
\end{align}
From here on we define gradient vectors $\nabla \mathcal{L}^{[l]}$ for each layer containing all partial derivatives in a single column by vectorization $\vect(\cdot)$ of their partial derivative matrix, which has the same dimensions as the respective weight matrix and bias vector.
\begin{align}
    \vect(\bm{A}) = 
    \begin{pmatrix}
    \vect(\frac{\partial \mathcal{L}}{\partial \bm{W}^{[l]}}) &
    \vect(\frac{\partial \mathcal{L}}{\partial \bm{b}^{[1]}}) 
    \end{pmatrix}^T
\end{align}
We define a gradient vector for each weight matrix of layer $l$ as:
\begin{align}
    \nabla \mathcal{L}^{[l]} = 
    \begin{pmatrix}
    \vect(\frac{\partial \mathcal{L}}{\partial \bm{W}^{[l]}}) &
    \vect(\frac{\partial \mathcal{L}}{\partial \bm{b}^{[1]}}) 
    \end{pmatrix}^T
\end{align}
Then, we concatenate all gradient vectors for use in residual weighting methods mentioned in later sections, which gives us the total gradient vector:
\begin{align}
    \nabla \mathcal{L} = 
    \begin{pmatrix}
    \nabla \mathcal{L}^{[1]} & 
    \nabla \mathcal{L}^{[2]} & 
    \hdots & 
    \nabla \mathcal{L}^{[L]} &
    \end{pmatrix}^T
\end{align}

\section{Developments}
\label{sec:Developments}

\subsection{General Remarks}
\label{sec:GeneralRemarks2}
The inclusion of sensitivities during training of neural networks provides performance improvement in the form of improved loss convergence, lower final loss and higher accuracy on validation data.
When having data provided by a numerical simulation such as FEM, which is capable of also providing sensitivities w.r.t. design parameters through adjoint or direct methods, Sobolev training is a quick and worthwhile adjustment.
Further in this section, Sobolev training will be expanded on, to provide an expansion of the basic ANN theory in the earlier section.
When considering this Sobolev training, a weighting issue arises, as the neural network training has to consider responses as well as sensitivities in its algorithm.
In regards to this issue, the first step is to find the optimal weights of said objectives for the optimal convergence performance.
This however begs the question of how to decide which of the responses and sensitivities are in how far important for the convergence performance and ultimately the accuracy of the final trained model.

\subsection{Sobolev Training}
\label{sec:SobolevTraining}
Sobolev trained neural networks follow the general structure of a basic ANN.
Instead of adding additional outputs by increasing the output layer size, the target loss function used for the optimization through gradient descent methods is adjusted to consider the additional sensitivities with respect to the inputs.
The equations will show the similarity of terms of partial derivatives inside the neural network, explaining why Sobolev training is computationally efficient, when these sensitivities are available.

\subsubsection{Preprocessing}
\label{sec:SobolevPreprocessing}
For the preprocessing, to obtain the sensitivity of the first standardised response to a standardised input, $\frac{\partial y_{1,s}}{\partial x_{e,s}}$, the corresponding standard deviations $\sigma_{x_e}, \sigma_{y_1}$ are applied to the sensitivity $\frac{\partial y_1}{\partial x_e}$ as follows:
\begin{align}
    \frac{\partial y_{1,s}}{\partial x_{e,s}} = \frac{\partial y_1}{\partial x_e} \cdot \frac{\sigma_{x_e}}{\sigma_{y_1}}
\end{align}
This is done for each element $e$ of $\bm{x}$ and for every element of $\bm{y}$.
Going forward, we leave out subscripts $d, e, s$ for simplicity.

\subsubsection{Sensitivities}
\label{sec:SobolevSensitivities}
When sensitivities of a system are available, especially when their additional calculation is computationally cheap, the implementation of sensitivities during training can effectively be considered.
The main goal of applying sensitivities for model training is to compare the sensitivity of the model response defined by its model parameters to the true sensitivity of the true response.
As a general chain rule expression, again for a neural network with only a single neuron in each of its layers, the sensitivity of the model $\frac{\partial \hat{y}}{\partial x}$ is defined for input $x$ and output $\hat{y}$:
\begin{align}
    \frac{\partial \hat{y}}{\partial x} = \frac{\partial \hat{y}}{\partial o^{[L]}} \frac{\partial o^{[L]}}{\partial z^{[L}} \frac{\partial z^{[L]}}{\partial o^{[L-1]}} \dots \frac{\partial o^{[2]}}{\partial z^{[2]}} \frac{\partial z^{[2]}}{\partial x}
\end{align}
With multiple inputs and outputs, we have the Jacobian $\frac{\partial \bm{\hat{y}}}{\partial \bm{x}} \in R^{\, n^{[L]} \times n^{[0]}}$ which is defined as:
\begin{align}
    \frac{\partial \bm{\hat{y}}}{\partial \bm{x}} = 
    \begin{bmatrix}
        \frac{\partial \hat{y}_1}{\partial x_1} & \dots & \frac{\partial \hat{y}_1}{\partial x_{n^{[0]}}} \\
        \vdots  & \ddots & \vdots \\
        \frac{\partial \hat{y}_{n^{[L]}}}{\partial x_1} & \dots & \frac{\partial \hat{y}_{n^{[L]}}}{\partial x_{n^{[0]}}}
    \end{bmatrix}
\end{align} 
When considering multiple neurons, we again introduce matrices and vectors per earlier definition.
However, we redefine $\bm{\delta}^{[L]}$, so as not to include the derivation of the loss function:
\begin{align}
    \bm{\delta}^{[L]} = \frac{\partial \bm{\hat{y}}}{\partial \bm{z}^{[L]}} \in R^{n^{[L]} \times 1}
\end{align}
We can now easily define the partial derivative with respect to inputs as follows:
\begin{align}
    \frac{\partial \bm{\hat{y}}}{\partial \bm{x}} = (\bm{W}^{[1]})^T \otimes \bm{\delta}^{[1]}
\end{align}
This contains all sensitivities of the neural network model.
All the partial derivatives for loss gradient calculations are already evaluated during the forward pass and backpropagation of the basic ANN.
This allows efficient computation of sensitivity related expressions without increasing training time comparatively to the basic neural network, emphasising the performance improvement through Sobolev training.
The new loss $\mathcal{L}$ is now defined as the sum of the response losses $\mathcal{E}_{\mathsf{Response}}$ and the sensitivity losses $\mathcal{E}_{\mathsf{Sensitivity}}$:
\begin{align}
    \mathcal{L} = \mathcal{E}_{\mathsf{Response}} + \mathcal{E}_{\mathsf{Sensitivity}}
\end{align}
Similar to the response losses $\mathcal{E}_{\mathsf{Response}}$, the sensitivity losses $\mathcal{E}_{\mathsf{Sensitivity}}$ are expressed as the half-mean squared error of the computed model's sensitivity and the true sensitivities.
For each input, the respective sensitivity loss is computed with the Euclidean vector norm of its vectors $\frac{\partial \bm{\hat{y}}}{\partial x_j}$ and $\frac{\partial \bm{y}}{\partial x_j}$:
\begin{align}
    \begin{split}
        \mathcal{E}_{\mathsf{Sensitivity}} &= \frac{1}{2} \cdot \sum_{j=1}^{n^{[0]}} \lVert\frac{\partial \bm{\hat{y}}}{\partial x_j} - \frac{\partial \bm{y}}{\partial x_j}\rVert^2 \\ 
        &= \frac{1}{2} \cdot \sum_{i=1}^{n^{[L]}} \sum_{j=1}^{n^{[0]}} \left(\frac{\partial  \hat{y}_i}{\partial x_j} - \frac{\partial y_i}{\partial x_j}\right)^2    
    \end{split}
\end{align}
The model training occurs unchanged, by solving the minimization of its defined loss $\mathcal{L}$ for its model parameters $\bm{\theta}$:
\begin{align}
    \begin{split}
        \bm{\theta} &= \arg \min_{\bm{\theta}} \mathcal{L}(\bm{\theta}) \\
        \textrm{with} \ \bm{\theta} &= \{\bm{W^{[l]}}, \bm{b^{[l]}}\} \ for \ l \in [1,L]
    \end{split}
\end{align}
While the sensitivities are expanding the loss function, they are not an output of additional neurons in the output layer.
The Sobolev training operates akin to a loss regularization term.
\subsubsection{Backpropagation}
\label{sec:SobolevBackpropragation}
Unchanged to the basic ANN, the process of computing the gradients with respect to the model parameters of layer $l = p$ is done by using automatic differentiation.
The chain rule is applied during backpropagation to evaluate each partial derivative step by step numerically. The general equation for partial derivatives $\frac{\partial \mathcal{L}}{\partial \bm{\theta}^{[p]}}$ of the loss $\mathcal{L}$ with respect to model parameters $\bm{\theta}^{[p]}$ follows, whereas an additional addend expands the previous equation, that is the partial derivative of the loss along the sensitivities, $\frac{\partial \mathcal{L}}{\partial \frac{\partial \bm{\hat{y}}}{\partial \bm{x}}} \frac{\partial \frac{\partial \bm{\hat{y}}}{\partial \bm{x}}}{\partial \theta^{[p]}}$:
\begin{align}
    \begin{split}
        \frac{\partial \mathcal{L}}{\partial \theta^{[p]}} &= \frac{\partial \mathcal{L}}{\partial \bm{\hat{y}}} \frac{\partial \bm{\hat{y}}}{\partial \bm{o}^{[L]}} \frac{\partial \bm{o}^{[L]}}{\partial \bm{z}^{[L]}} \frac{\partial \bm{z}^{[L]}}{\partial \bm{o}^{[L-1]}} \dots \frac{\partial \bm{o}^{[p]}}{\partial \bm{z}^{[p]}} \frac{\partial \bm{z}^{[p]}}{\partial \theta^{[p]}} \\ 
        &+ \frac{\partial \mathcal{L}}{\partial \frac{\partial \bm{\hat{y}}}{\partial \bm{x}}} \frac{\partial \frac{\partial \bm{\hat{y}}}{\partial \bm{x}}}{\partial \theta^{[p]}} \ ,\ with \ \ 1 \leq p \leq L
    \end{split}
\end{align}
For $\frac{\partial \mathcal{L}}{\partial \frac{\partial \bm{\hat{y}}}{\partial \bm{x}}}$, the evaluation is simply the difference between the neural networks sensitivity and the true sensitivity:
\begin{align}
    \frac{\partial \mathcal{L}}{\partial \frac{\partial \bm{\hat{y}}}{\partial \bm{x}}} = \frac{\partial \bm{\hat{y}}}{\partial \bm{x}} - \frac{\partial \bm{y}}{\partial \bm{x}}
\end{align}
As previously shown, the sensitivity of the model response is also expressed per chain rule.
Therefore, for layer $l = p$, it simply follows for $\frac{\partial }{\partial \theta^{[p]}} \cdot \left(\frac{\partial \bm{\hat{y}}}{\partial \bm{x}}\right)$:
\begin{align}
    \begin{split}
        \frac{\partial \left(\frac{\partial \bm{\hat{y}}}{\partial \bm{x}}\right)}{\partial \theta^{[p]}} 
        = \frac{\partial }{\partial \theta^{[p]}} \cdot &\bigg( \frac{\partial \bm{\hat{y}}}{\partial \bm{o}^{[L]}} \frac{\partial \bm{o}^{[L]}}{\partial \bm{z}^{[L]}} \frac{\partial \bm{z}^{[L]}}{\partial \bm{o}^{[L-1]}} \\ 
        &\dots \frac{\partial \bm{o}^{[1]}}{\partial \bm{z}^{[1]}} \frac{\partial \bm{z}^{[1]}}{\partial \bm{x}} \bigg)
    \end{split}
\end{align}
For the partial derivative relating to the model sensitivity, that is the partial derivative with respect to input, for layer $p$ this expression has to be differentiated for all layers $l \geq p$.
This is because $\theta^{[p]}$ is nested inside all layers after layer $p$ propagation-wise.
Consider the last layer for our simple expressions of depth $1$ in all layers, with the previous defined state of all parts of $\frac{\partial }{\partial \theta^{[p]}} \cdot \left(\frac{\partial \bm{\hat{y}}}{\partial \bm{x}}\right)$, for the case $p = L$, this expression turns to:
\begin{align}
    \frac{\partial \left(\frac{\partial \bm{\hat{y}}}{\partial \bm{x}}\right)}{\partial \theta^{[L]}}
    = \frac {\partial \left(\frac{\partial \bm{\hat{y}}}{\partial \bm{o}^{[L]}} \frac{\partial \bm{o}^{[L]}}{\partial \bm{z}^{[L]}} \frac{\partial \bm{z}^{[L]}}{\partial \bm{o}^{[L-1]}}\right)} {\partial \theta^{[L]}} \cdot \left(\dots \frac{\partial \bm{o}^{[1]}}{\partial \bm{z}^{[1]}} \frac{\partial \bm{z}^{[1]}}{\partial \bm{x}}\right)
\end{align}
Since the goal is to compute the partial derivative with respect to $\theta^{[L]}$, any terms propagated before this model parameter are independent of it, allowing these independent terms to be factored out of the partial derivative with respect to $\theta^{[L]}$. The partial derivatives of $\frac {\partial \left(\frac{\partial \bm{\hat{y}}}{\partial \bm{o}^{[L]}} \frac{\partial \bm{o}^{[L]}}{\partial \bm{z}^{[L]}} \frac{\partial \bm{z}^{[L]}}{\partial \bm{o}^{[L-1]}}\right)} {\partial \theta^{[L]}}$ are left for differentiation, which were previously defined.
Per product and chain rule, it follows:
\begin{align}
    \frac {\partial \left(\frac{\partial \bm{\hat{y}}}{\partial \bm{o}^{[L]}} \frac{\partial \bm{o}^{[L]}}{\partial \bm{z}^{[L]}} \frac{\partial \bm{z}^{[L]}}{\partial \bm{o}^{[L-1]}}\right)} {\partial \theta^{[L]}} = \frac {\partial \left( 1 \cdot a'(z^{[L]}) \cdot w^{[L]}\right)} {\partial \theta^{[L]}}
\end{align}
and for $\theta$ being a weight $w$ and bias $b$, respectively:
\begin{align}
    \frac {\partial \left( 1 \cdot a'(z^{[L]}) \cdot w^{[L]}\right)} {\partial \theta^{[L]}} = \begin{cases}
        \begin{aligned}
            &a''(z^{[L]}) \cdot o^{[L-1]} \cdot w^{[L]} +  a'(z^{[L]}) \\ 
            &\text{for  } \theta^{[L]} = w^{[L]} \\ \\
            &a''(z^{[L]}) \cdot w^{[L]} \\ 
            &\text{for  } \theta^{[L]} = b^{[L]}        
        \end{aligned}
    \end{cases}
\end{align}
Through these general equations, all gradients for the Sobolev trained neural network can be computed.
In the case of a simple activation function, such as the rectified linear unit applied in this work, second-order derivatives of the activation function are $0$, simplifying the gradient calculation.
When considering multiple neurons, with the previous definition of $\frac{\partial  \bm{\hat{y}}}{\partial \bm{x}}$ and our newly defined loss function, we also define the first partial derivative with respect to model parameters for the sensitivities:
\begin{align}
     \frac{\partial \left( 
     \frac{\partial  \bm{\hat{y}}}{\partial \bm{x}} \right)
     }{\partial \bm{W^{[1]}}}  =
     \bm{I}^{W^{[1]}}
     \otimes
     \bm{\delta}^{[1]}
\end{align}
With $\bm{I}^{W^{[1]}}$ being an Identity matrix with the same dimensions as its respective weight matrix of the first layer.
For all other layers, we have:
\begin{align}
     \frac{\partial \left( 
     \frac{\partial  \bm{\hat{y}}}{\partial \bm{x}} \right)
     }{\partial \bm{W^{[l]}}}  =
     \frac{\partial \left( \bm{W}^{[1]}
     \otimes
     \bm{\delta}^{[1]}\right)}{\partial \bm{W^{[l]}}}
\end{align}
Following the chain rule, when reaching the respective layer $\bm{\delta}^{[l-1]}$, we obtain:
\begin{align}
     \frac{\partial \bm{\delta}^{[1-1]}}{\partial \bm{W^{[l]}}} =
     \bm{I}^{W^{[l-1]}}
     \otimes
     \bm{\delta}^{[l]}
     \odot
     \frac{\partial \bm{o}^{[l]}}{\partial \bm{z}^{[l]}}
\end{align}
Since we use the ReLU activation function, all partial derivatives per product rule of $\frac{\partial \bm{o}^{[l]}}{\partial \bm{z}^{[l]}}$ turn $0$ and are left out of this equation.
The gradients of the Sobolev loss are now defined as:
\begin{align}
     \nabla \mathcal{L} = \frac{\partial \mathcal{L}}{\partial \bm{\hat{y}}} \frac{\partial \bm{\hat{y}}}{\partial \bm{W^{[l]}}} 
     +
     \frac{\partial \mathcal{L}}{\partial \left(\frac{\partial \bm{\hat{y}}}{\partial \bm{x}}\right)} 
     \frac {\partial \left(\frac{\partial \bm{\hat{y}}}{\partial \bm{x}}\right)}{\partial \bm{W^{[l]}}}
\end{align}
Once all gradients have been computed, the applied training algorithm adjusts each model parameter unchanged to the basic ANN by an update step which now considers the gradients w.r.t. model parameters for both responses as well as sensitivities.

\subsection{Residual Weights}
\label{sec:ResidualWeights}
When considering the different losses of response and sensitivity, the optimization task has become more involved.
Each individual loss represents its own optimization target for which convergence towards zero will improve accuracy of the neural network model.
But since the actual target value of the model is to approximate the true output $y$ only, it is still considered a single-objective optimization problem.
Nevertheless, by introducing multiple new losses, which are indirectly approximated, a new space of loss optimization can be considered.
In this more involved optimization problem, there is an optimal weighting of response and sensitivity losses to obtain optimal training convergence.
When considering the optimal ratio of the loss addends, there are a variety of approaches to consider.
In this work, various methods to optimise the residual weights are considered and empirically evaluated.
All optimization methods used for the residual weights of the losses will apply the same optimization algorithm, ADAM, which is applied to the model parameters of the neural network.
The general goal of these approaches is, that during any training step, the defined sum of response and sensitivity losses arrange a traversable space by ratio of residual weights, where there should exist a ratio for optimal performance or convergence during its training.
\begin{figure}[ht]
\centering
\includegraphics[scale=0.3]{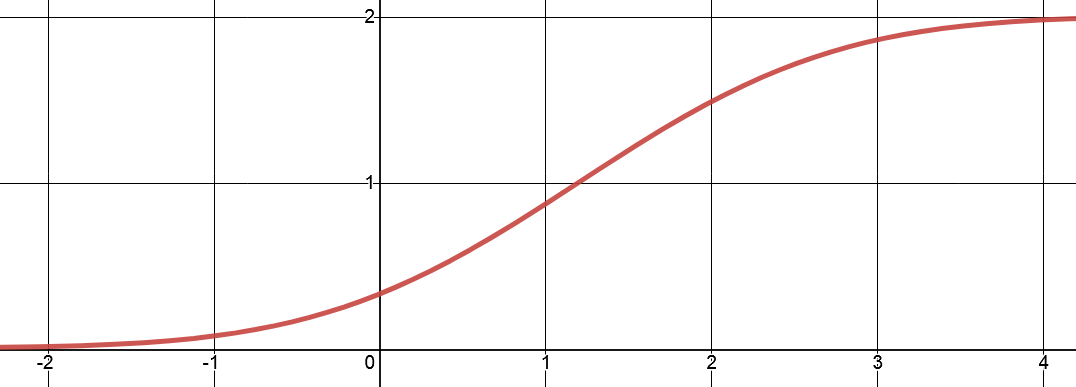}
\caption{Residual weight function}
\label{fig:errorfnc}
\end{figure}
The methods are employed to find this ratio and thereby improve convergence.
To make this possible, the previous loss $\mathcal{L}$ is expanded by residual weights $\lambda_R$ and $\lambda_S$:
\begin{align}
    \mathcal{L} = \lambda_R \ \mathcal{E}_{\mathsf{Response}} + \lambda_S \ \mathcal{E}_{\mathsf{Sensitivity}}
\end{align}
To limit the range of the residual weights $\lambda_i$ and stabilise the training process, they are evaluated by a function.
This limits the residual weights between the range $[\epsilon_0;2]$ and $\epsilon_0 = 0.01$ is added to avoid elimination of response and sensitivity loss and numerical instability.
Specifically, the initial residual weight $\lambda_{i,prev}$ is applied to the error function $\erf(\cdot)$, which returns $\lambda_i$, which is then used for subsequent computation:
\begin{align}
    \lambda_i = 1 - \erf \left(\frac{1.2-\lambda_{i,prev}}{\sqrt{3}}\right) + \epsilon_{0} 
\label{eq:errorfnc}
\end{align}
This is then repeated for during every training step to reevaluate the residual weights.
While previously, the loss was minimized for model parameters $\bm{\theta}$, now the model training occurs by solving the defined loss $\mathcal{L}$ twofold.
First, the loss $\mathcal{L}$ is minimized for its model parameters $\bm{\theta}$, just like with a regular neural network:
\begin{align}
    \begin{split}
        \bm{\theta} &= \arg \min_{\bm{\theta}} \mathcal{L}(\bm{\theta, \lambda}) \\ \textrm{with} \ \bm{\theta} &= \{\bm{W^{[l]}}, \bm{b^{[l]}}\} \ \textrm{for} \ l \in [1,L]        
    \end{split}
\end{align}
Lastly, the various methods listed in table \ref{table:modes} are used to update the residual weights $\lambda_i$.
These methods cover a variety of optimization targets during training.
Certain methods only differ in their application of said optimization target $\mathcal{G}(\cdot)$:
\begin{align}
    \begin{split}
        \bm{\lambda} &= \arg \min_{\bm{\lambda}} \mathcal{G}(\bm{\theta, \lambda}) \\ 
        \textrm{with} \ \bm{\lambda} &= [\epsilon_0;2+\epsilon_0] \ \textrm{for} \ \epsilon_0 = 0.01
    \end{split}
\end{align}
Table \ref{table:modes} covers all applied methods for the optimization of the residual weights $\bm{\lambda}$ inside the loss function.
These methods only apply to the residual weight optimization, in parallel there is still the usual optimization of neural network weights and biases $\bm{\theta}$ per minimization of the MSE loss.
In order they are 1, minimization and 2, maximization of the total loss function $\mathcal{L}$.
These are the most basic options of target functions, whereby minimising the loss follows the same idea as the optimization of model parameters.
The maximization of loss w.r.t. residual weights $\bm{\lambda}$ in parallel to the general minimization of the loss w.r.t. model parameters $\theta$ should allow for the maximum possible gradient step correction.
Arguably, it is also expected that minimization of the loss will simply converge the residual weights to $0$ and the maximization of the loss will lead to increasing residual weights since losses are defined as $>0$.
Then we have for 3, minimization and 4, maximization of the magnitude of the total loss gradient $\lvert \nabla\mathcal{L} \rvert$.
These target functions directly optimise in regards to the convergence rate $\lvert \nabla\mathcal{L} \rvert$.
The idea and expected behaviors are similar to the methods 1 and 2.
Next, for method 5, minimization of the variance between $Var(\mathcal{L})$ each individual loss, which aims to reduce the difference in performance between the individual losses.
This should lead to all individual losses working in tandem, either all performing well or all performing bad.
Methods 6 to 9 are various methods targeting the angle between different loss gradient vectors through the cosine similarity.
Cosine losses or methods of targeting cosine similarity in neural networks are used in various works, most notably in image recognition, where similar feature vectors are compared \cite{NEURIPS2021_cbcb58ac,10.1007/978-3-642-19309-5_55, Barz_2020_WACV}.
In the case of 6 and 7, the target is to align the gradient vector of each individual loss $\nabla\mathcal{L}_i$ to the total loss gradient vector $\nabla\mathcal{L}$ by minimising the cosine distance and the squared cosine similarity, respectively.
Similarly, methods 8 and 9 apply the minimization of the cosine distance and the squared cosine similarity.
However, in their case, the target is to align each individual loss gradient vector $\nabla\mathcal{L}_i$ with the gradient vector of the loss for the response $\nabla\mathcal{L}_R$ instead, since the response is the main target of the model.
The cosine distance minimization should force equal angles between the respective gradient vectors.
Similarly, the squared cosine distance minimization should force orthogonality between the respective gradient vectors.

\begin{table}[ht]
\centering
\caption{Methods for optimising the residual weights $\lambda$}
\renewcommand{\arraystretch}{1.2}
\begin{tabular}{|c | c | c|} 
 \hline
 \# & Method\\
 \hline\hline
 1 & $\arg \min_{\bm{\lambda}} \mathcal{L}$\\
 \hline
 2 & $\arg \max_{\bm{\lambda}} \mathcal{L}$\\ 
 \hline
 3 & $\arg \min_{\bm{\lambda}} \lvert \nabla\mathcal{L} \rvert$\\ 
 \hline
 4 & $\arg \max_{\bm{\lambda}} \lvert \nabla\mathcal{L} \rvert$\\
 \hline
 5 & $\arg \min_{\bm{\lambda}} Var(\mathcal{L})$\\
 \hline
 6 & $\arg \min_{\bm{\lambda}} \left(1 - \frac{\nabla \mathcal{L} \cdot \nabla \mathcal{L}_i}{\lvert \nabla \mathcal{L} \rvert \cdot \lvert \nabla \mathcal{L}_i \rvert}\right)$\rule{0pt}{3ex}\rule[-1.8ex]{0pt}{0pt}\\
 \hline
 7 & $\arg \min_{\bm{\lambda}} \left(\frac{\nabla \mathcal{L} \cdot \nabla \mathcal{L}_i}{\lvert \nabla \mathcal{L} \rvert \cdot \lvert \nabla \mathcal{L}_i \rvert}\right)^2$\rule{0pt}{3.4ex}\rule[-1.8ex]{0pt}{0pt}\\
 \hline
 8 & $\arg \min_{\bm{\lambda}} \left(1- \frac{\nabla \mathcal{L}_{\mathsf{Response}} \cdot \nabla \mathcal{L}_j}{\lvert \nabla \mathcal{L}_{\mathsf{Response}} \rvert \cdot \lvert \nabla \mathcal{L}_j \rvert}\right)$\rule{0pt}{3ex}\rule[-1.8ex]{0pt}{0pt}\\
 \hline
 9 & $\arg \min_{\bm{\lambda}} \left(\frac{\nabla \mathcal{L}_{\mathsf{Response}} \cdot \nabla \mathcal{L}_j}{\lvert \nabla \mathcal{L}_{\mathsf{Response}} \rvert \cdot \lvert \nabla \mathcal{L}_j \rvert} \right)^2$\rule{0pt}{3.4ex}\rule[-1.8ex]{0pt}{0pt}\\
 \hline
\end{tabular}
\end{table}

For comparison, the results will include additional model versions, see Table \ref{table:modes_basic}, which omit an adaptive weighting scheme and the error function in (\ref{eq:errorfnc}).
The last column shows the initialised residual weight values $\lambda_{\mathsf{Initial State}}$, the first value weighting the response loss, the second value weighting the sensitivity w.r.t. the first input variable and the third value weighting the sensitivity w.r.t the second input variable.
Modes 10 and 11 are the Sobolev trained model and a basic ANN respectively, with the basic ANN not considering sensitivities and therefore setting the respective weights to zero.
In SNN the weights are fixed at one each, in order to have a baseline Sobolev for comparison, which we seek to improve.
There is also modes 12 and 13, which add exponential decay, see (\ref{eq:expDec}), and exponential increase, see (\ref{eq:expInc}), of the residual weight value.
The residual weights $\lambda_i$ are adjusted each epoch by the rate $\mu$, the current number of epochs $i_{Epochs}$ and the residual weight of the previous epoch $\lambda_{i,previous}$.

\label{table:modes}
\begin{align}
    \lambda_i = \frac{\lambda_{i,previous}}{1 + (\mu*i_{Epochs})}
\label{eq:expDec}
\end{align}
\begin{align}
    \lambda_i = \lambda_{i,previous}*(1 + (\mu*i_{Epochs}))
\label{eq:expInc}
\end{align}
\begin{table}[ht]
\centering
\caption{Methods with none or fixed residual weight optimization}
\renewcommand{\arraystretch}{1.2}
\begin{tabular}{|c | c | c|} 
 \hline
 \# & Method & $\lambda_{\mathsf{Initial State}}$ \\ [0.5ex] 
 \hline\hline
 10 & SNN & $[1,1,1]$\\ [0.5ex] 
 \hline
 11 & Basic ANN & $[1,0,0]$\\ [0.5ex] 
 \hline
 12 & SNN Exp. Decay & $[1,1,1]$\\ [0.5ex] 
 \hline
 13 & SNN Exp. Increase & $[0,0,0]$\\ [0.5ex] 
 \hline
\end{tabular}
\label{table:modes_basic}
\end{table}

\section{Tools and Application}
\label{sec:ToolsandApplication}

\subsection{General Remarks}
\label{sec:GeneralRemarks3}
For the realization of this work, a FEM simulation was used to provide the necessary training data to train multiple neural network models.
In addition, the following section will go into the details of each tool used.
\subsection{Finite Element Model}
\label{sec:FiniteElementModel}
\begin{figure}[ht]
\centering
\includegraphics[scale=0.3]{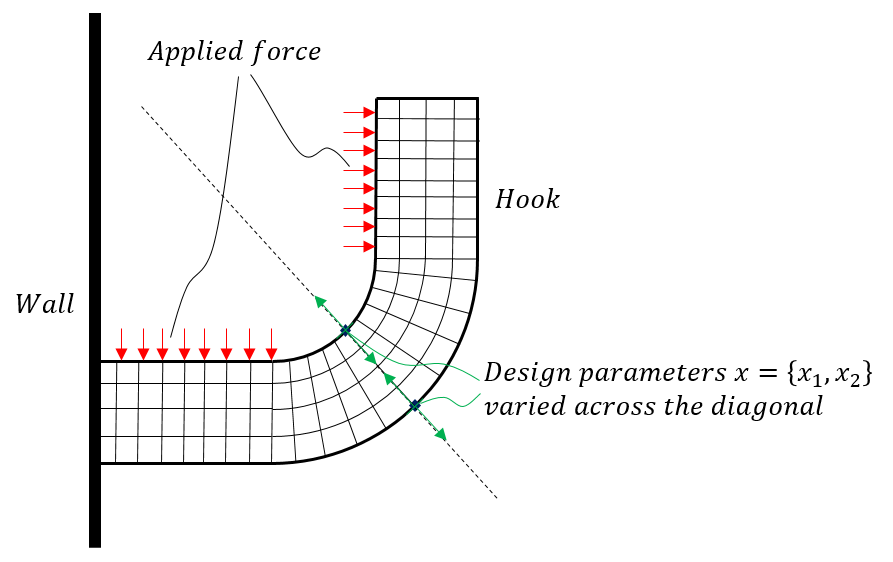}
\caption{Sketch of the hook system approximated by the finite element model. The design parameters $x_1,x_2$ in the range $[-1,1]$ adjust the blue marked mesh nodes along the green axis}
\label{fig:hooksketch}
\end{figure}

All neural network models approximate the data generated by a finite element model.
The FEM evaluates a two dimensional mechanical system of linear and non linear elasticity.
In this example, a simple hook, see figure \ref{fig:hooksketch}, is attached to a wall.
Further, a force is applied on top of the hook, to emulate an object being attached to the hook.
The shape of this hook is then modified through two design parameters.

\begin{figure}[ht]
\centering
\includegraphics[width=0.2\textwidth]{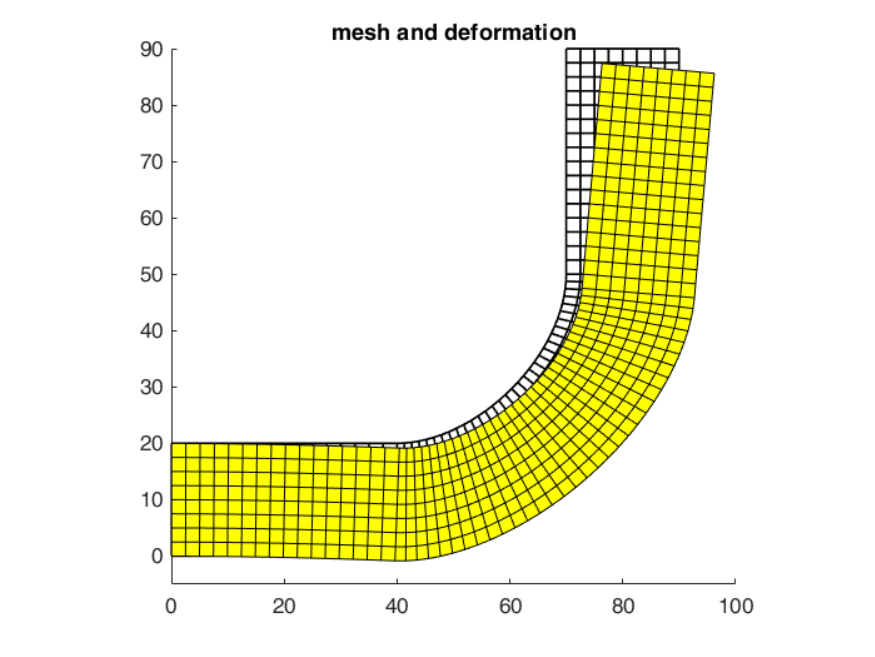}
\includegraphics[width=0.2\textwidth]{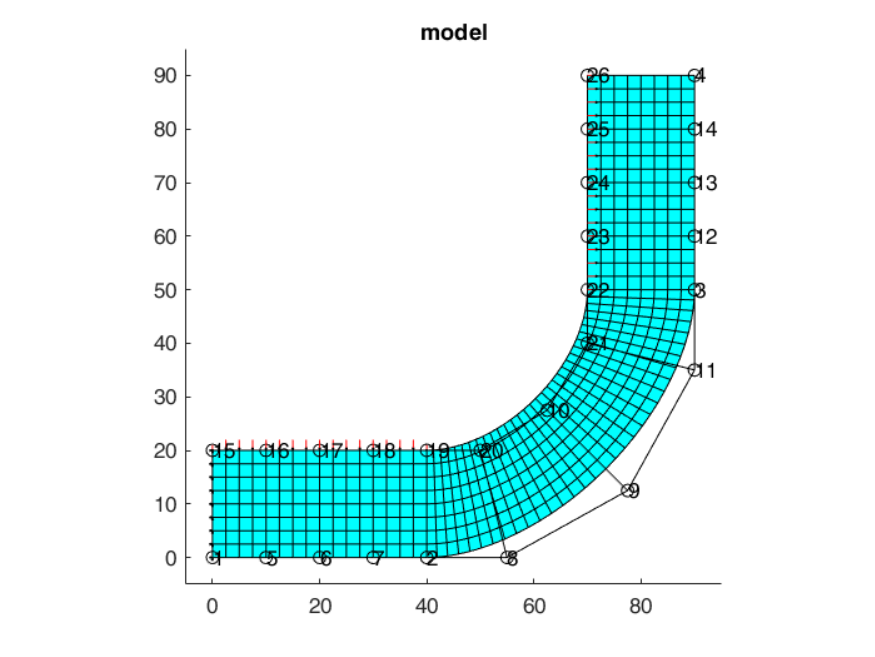}
\includegraphics[width=0.2\textwidth]{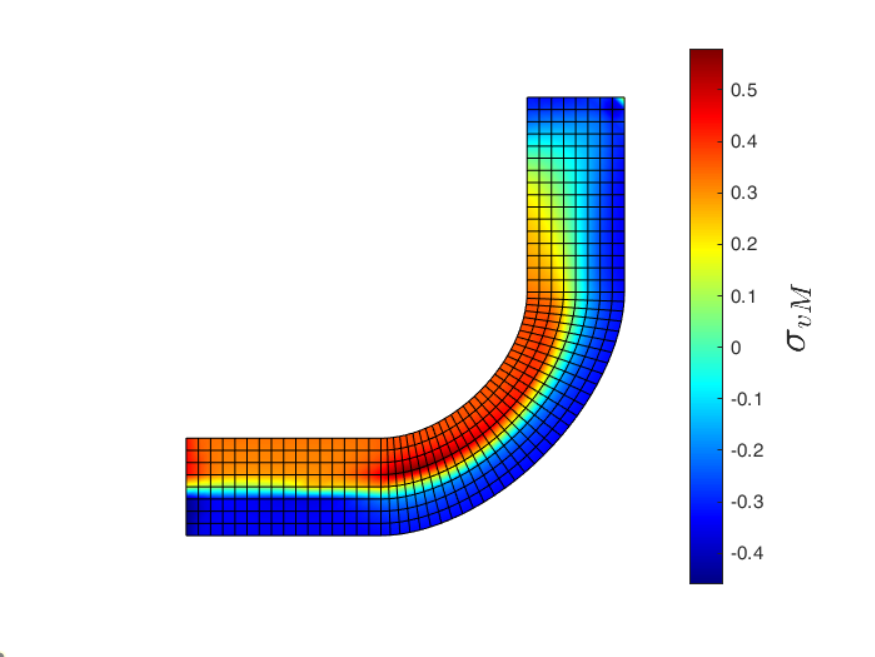}
\includegraphics[width=0.2\textwidth]{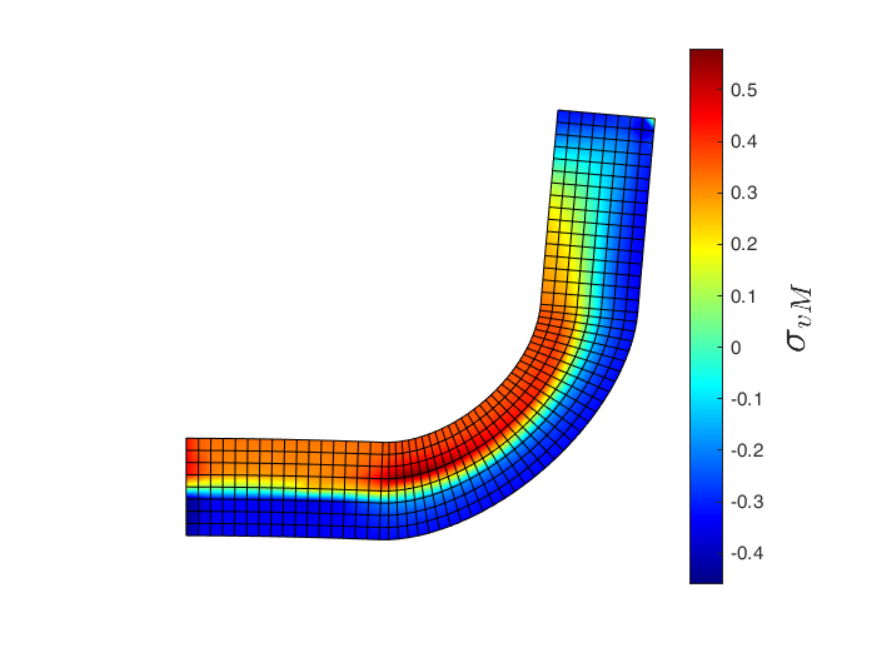}
\caption{Visualization of the hook; Top right: Control points; Top left: Mesh deformation; Bottom: Stress state for design parameters $x_1,x_2 = (0,0)$, non- and deformed.}
\label{fig:hookvisuals}
\end{figure}

Their values shift two controls points of an underlying geometry description in the midsection of the hook orthogonal to the hook radius, thereby reshaping the hook, see control points 9 and 10 marked in Figure \ref{fig:hookvisuals}.
The finite element model evaluates this mechanical system in terms of its stress state for an array of design parameter pairs.
The design parameters are evaluated in the range $[-1, 1]$ to stay in the range of physically plausible deformation of the hook geometry.
These responses and sensitivities are then used for training the neural network models.
The focus of this paper is the stress triaxility of the material.
Since the goal is to compare different residual weighting methods, the neural network design is simplified.
More precisely, the stress triaxility of a single specific node is chosen, where the finite element model evaluation produces the most critical value.
For the non linear elasticity evaluation, the material is specified as St. Venant-Kirchoff material and the finite element model evaluates the hook in the plane stress state.
For a general visualization of the use case, see figure \ref{fig:hookvisuals}.

\subsection{Neural Network Models}
\label{sec:NeuralNetworkModels}
\begin{figure}[ht]
\centering
\includegraphics[scale=0.25]{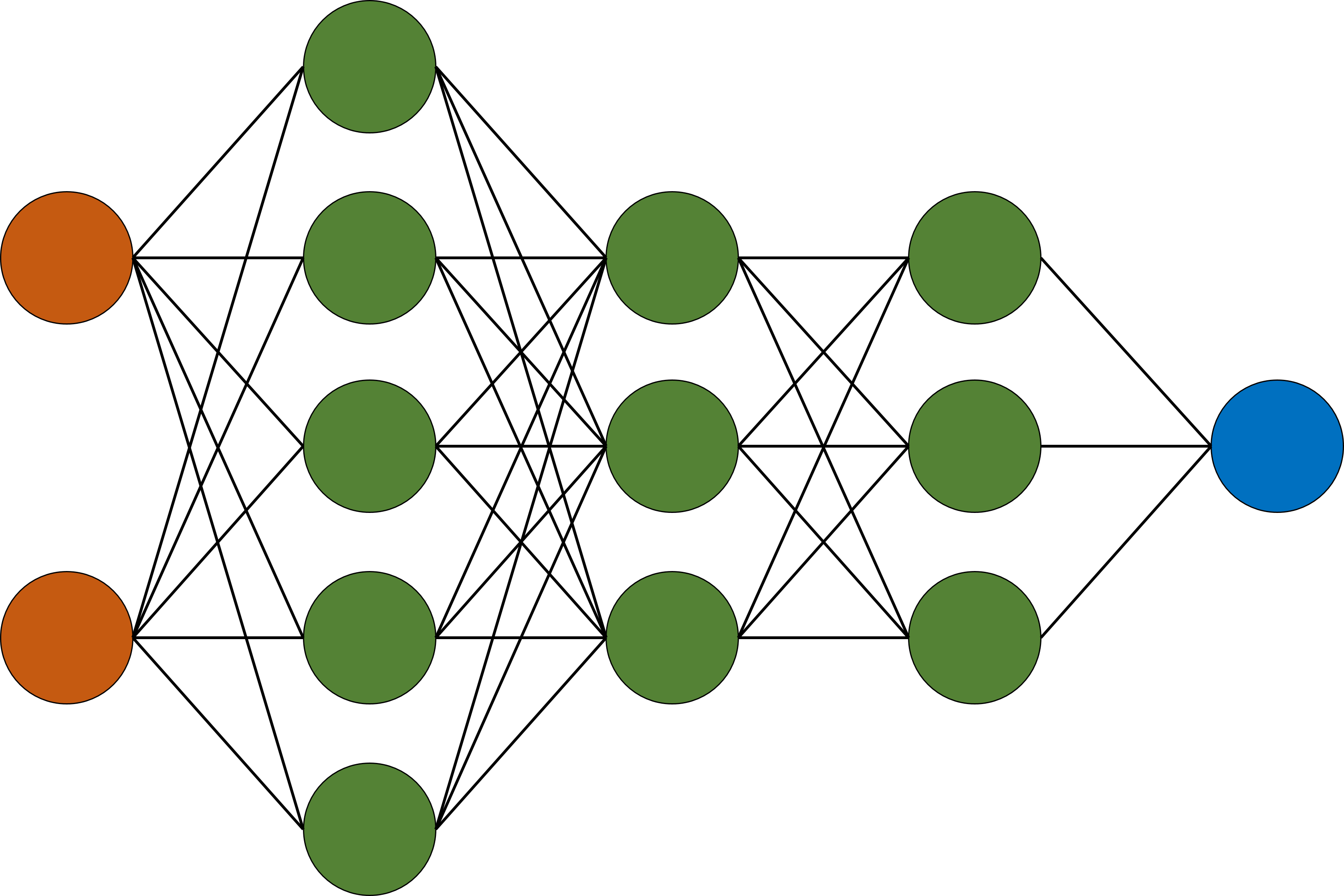} \\
\includegraphics[scale=0.25]{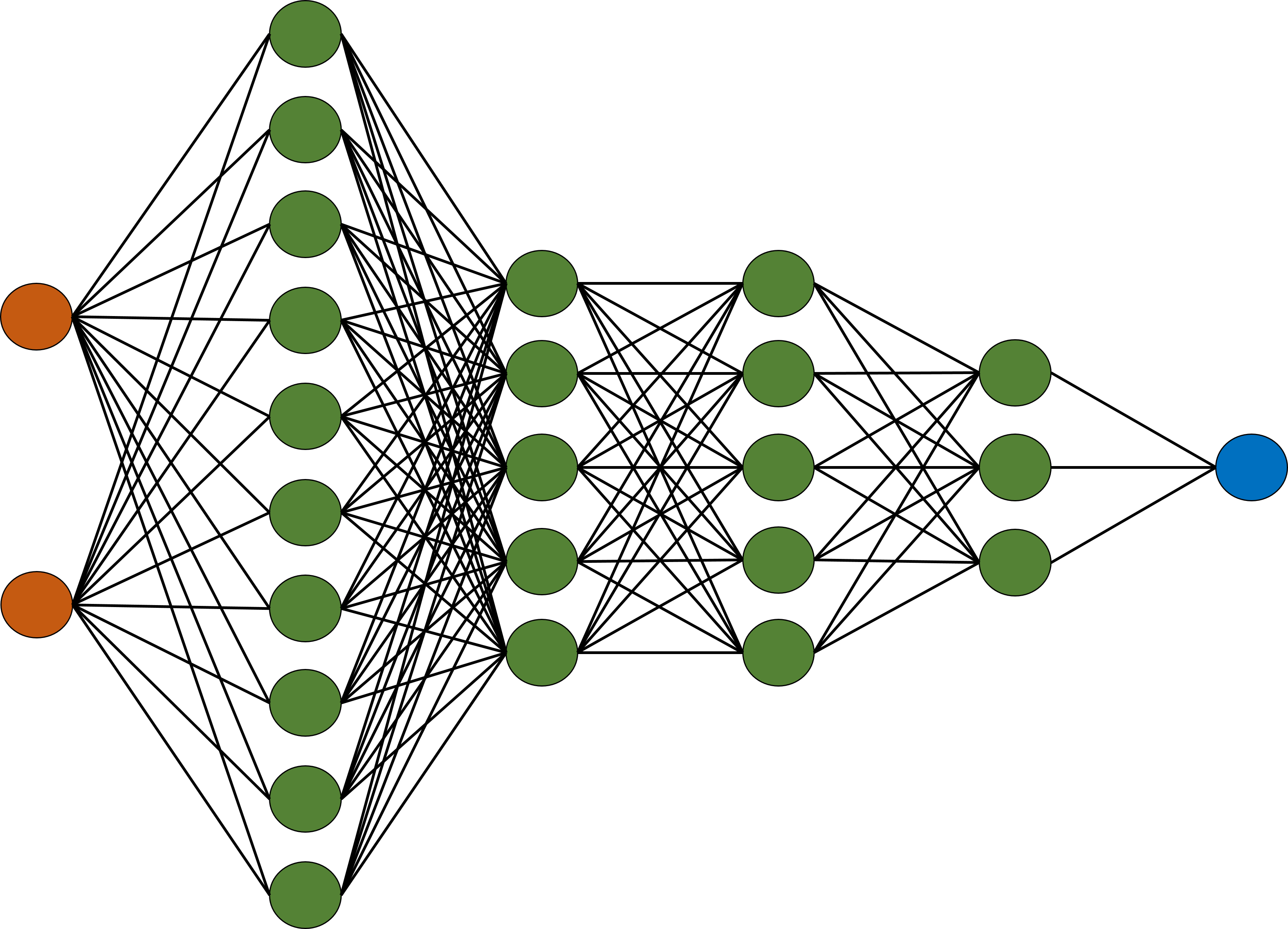}
\caption{Neural network model layer and neuron structure with 2 Inputs and 1 Output. Left: Basic layer structure 5-3-3, Right: Additional layer structure 10-5-5-3}
\label{fig:layer1and2}
\end{figure}

In this paper, all neural network models are trained in the MATLAB R2023b software environment.
To train the models, the MATLAB Deep Learning Toolbox is utilised.
The models are designed in small dimensions, with the hidden layers being made up of a "5-3-3" neuron-layer structure, see figure \ref{fig:layer1and2}.

Each model has two input layers and a single output layer.
Each layer applies the rectified linear unit as an activation function.
A second model layer structure is used later to analyse the effect of layer structure on training convergence.
The second structure will have an additional hidden layer and contain more neurons as "10-5-5-3", see figure \ref{fig:layer1and2}.
Furthermore, the training occurs in adequate minibatches using the ADAM training algorithm.
Each NN model is trained on the same dataset and each result is the average of 100 trained models with random initialization of model parameters, to reduce bias from various factors.
\begin{table}[ht]
\centering
\caption{Neural network hyperparameters}
\renewcommand{\arraystretch}{1.2}
\begin{tabular}{|c | c|}
 \hline
Hyperparameter & Value\\ [0.1ex] 
 \hline\hline
 Minibatches & $64$\\ 
 \hline
 Learnrate & $0.001$\\ 
 \hline
 Grad Decay & $0.9$\\ 
 \hline
 Squared Grad Decay & $0.999$\\
 \hline
 Epsilon & $10^{-8}$\\
 \hline
\end{tabular}
 \label{table:hyperparameters}
\end{table}
The model uses 320 training data points in addition to 305 validation data points, whereby 625 data points are generated by the FEM.
The data 625 points are evaluated in a range of design parameters of constant step size.
For the validation data, data points of uniform step size are extracted.
For each iteration, the current model state is used to calculate the relative $l_2$ error through use of the validation data.
Table \ref{table:hyperparameters} covers the important hyperparameters used to train the neural network models with the ADAM algorithm.
The first results are produced after only 500 epochs of training to gauge the effectiveness of each method.
Afterwards, additional results of the two best converging methods as well as the SNN are produced, which entails an expanded training for 1000 epochs.
There are also brief results for 2000 epochs.

\section{Results}
\label{sec:Results}
The results consist of multiple instances of the aforementioned models with varying training parameters.
While there are multiple metrics that can be observed, the focus will be on the $l_2$ error concerning the validation data.
This information is visualised through various tables and graphs.
Since the models use Sobolev training and adjust the magnitude of the individual loss addends, the loss value of the different models does not provide meaningful information for comparison.

\begin{figure}[ht]
\centering
\includegraphics[scale=0.55]{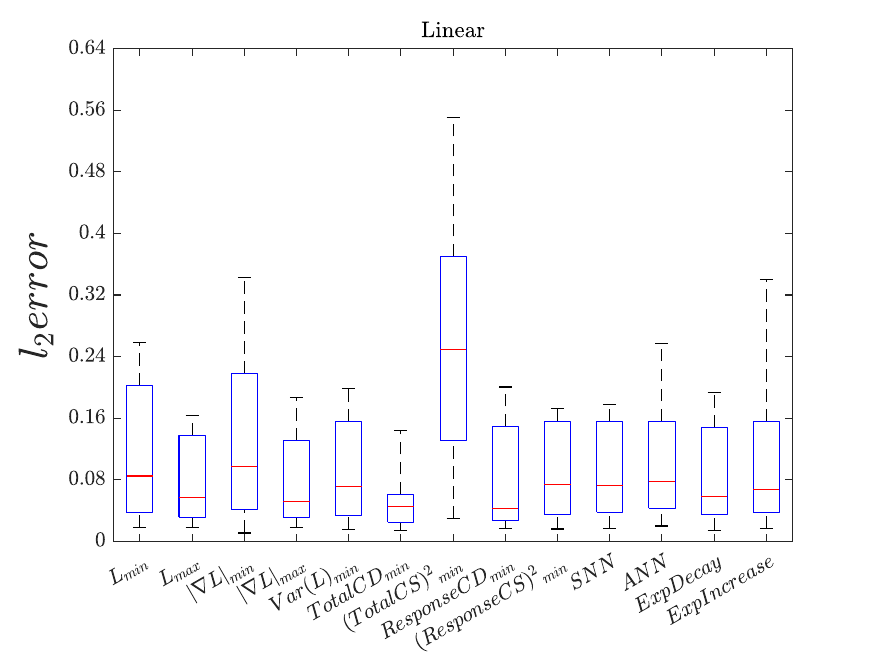}
\caption{$L_2$ error spread of all training runs for each mode}
\label{fig:500BestL2}
\end{figure}
Furthermore, out of all the trained models for each mode, the best performing models are compared via box plots, with a secondary focus on the best performing residual weighting methods.
The box plots visualise the average value over 100 trained models with a red line, with the inner $50\%$ of values of the trained models inside of the blue box.
The dotted black lines represent the rest of the trained models and the lower and upper end of the values.
Finally, for the best performing residual weighting methods, the 1000 epoch training runs are visualised with their loss and $l_2$ error curves over the iterations.
On top of this, the individual losses of responses and sensitivities, as well as the residual weights, are plotted, to analyse for possible patterns.
There are additional results for the purpose of analysing the effects of layers, epochs and activation functions, etc. on the training convergence of the models.

\subsection{Linear Case}
\label{sec:Linear Case}
\begin{figure}[ht]
\centering
\includegraphics[scale=0.5]{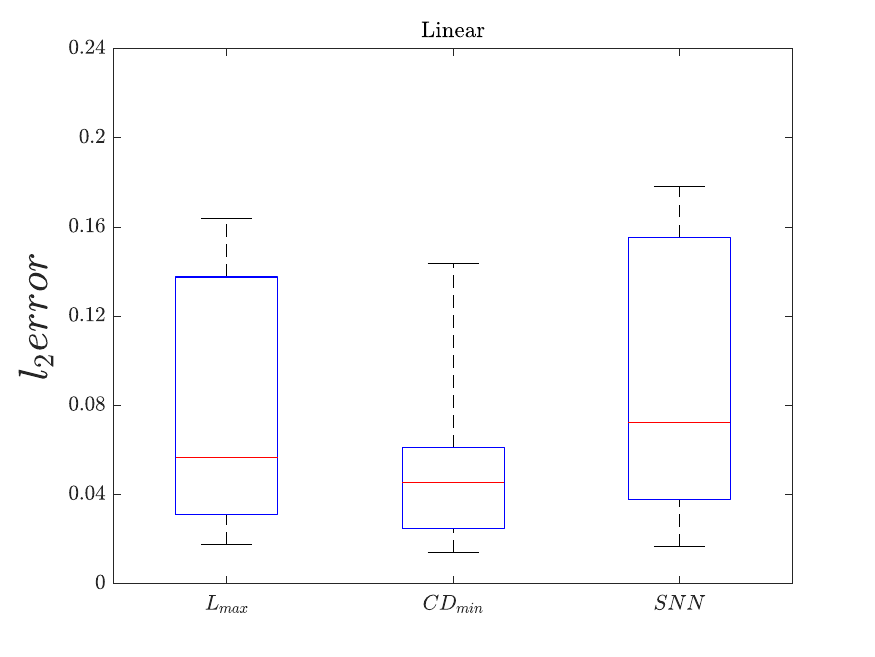}
\caption{$L_2$ error spread of $max \ \mathcal{L}$ and $min \ \bm{CD}$. For comparison the mode without residual weight optimization is included}
\label{fig:500L2}
\end{figure}

For the linear elasticity case, training the models for 500 epochs results in the box plots in figure \ref{fig:500BestL2} for the $l_2$ error values.
For better analysis, the $l_2$ error box-plots of residual weighting mode 2 and 6 are represented together with mode 9 for comparison in figure \ref{fig:500L2}.
In addition, the averaged $l_2$ error and training time results  of the multiple training runs for each mode are listed in table \ref{table:500avgL2}.

\begin{figure}[ht]
\centering
\includegraphics[scale=0.55]{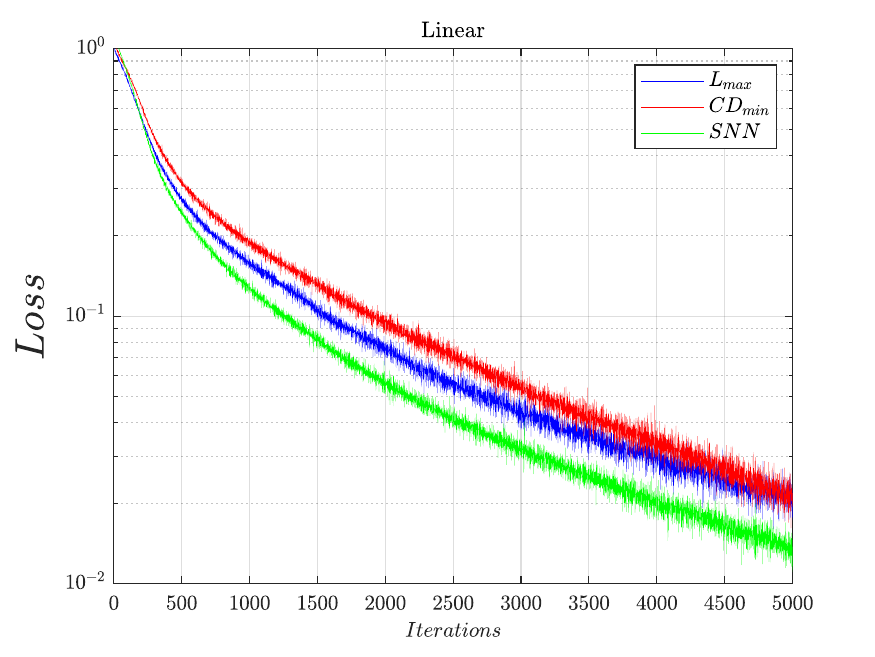} \
\includegraphics[scale=0.55]{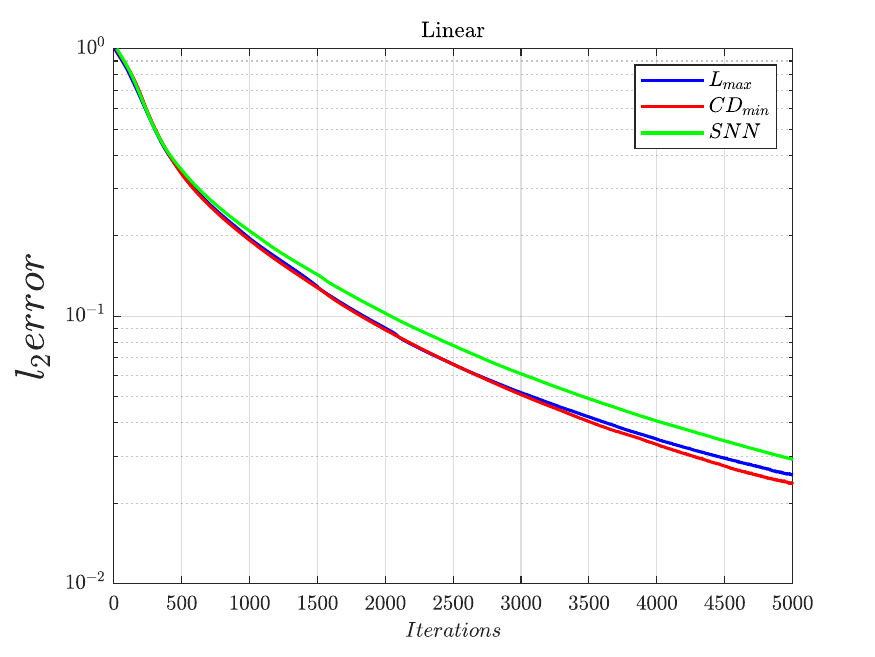}
\caption{\textit{Top}: Loss over the training iterations of $max \ \mathcal{L}$ and $min \ \bm{CD}$. For comparison the mode without residual weight optimization is included; Bottom: $L_2$ error over the training iterations of $max \ \mathcal{L}$ and $min \ \bm{CD}$. For comparison the mode without residual weight optimization is included}
\label{fig:1000LossandL2}
\end{figure}

\begin{figure}[ht]
\centering
\includegraphics[scale=0.55]{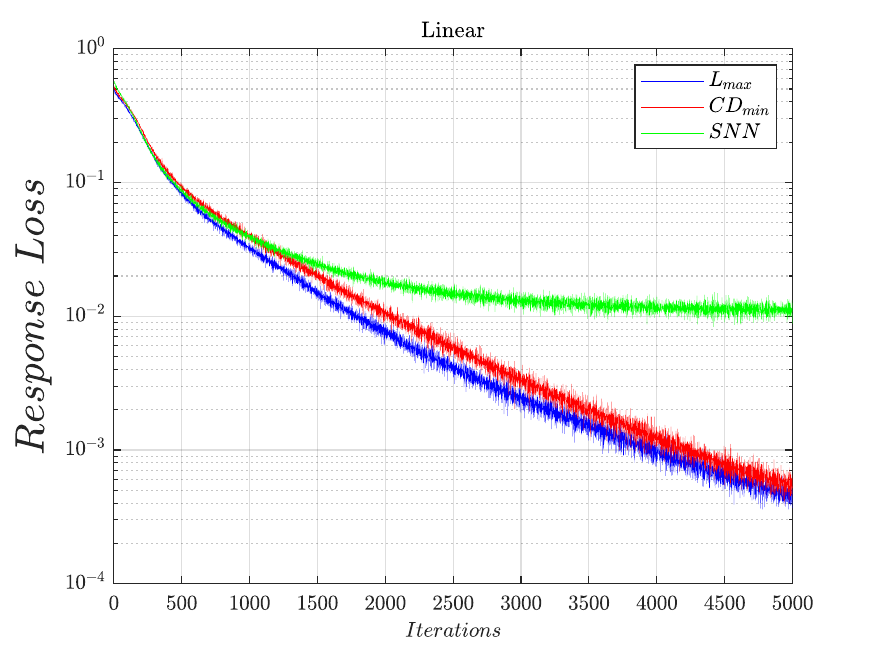}
\caption{Response loss over the training iterations of $max \ \mathcal{L}$ and $min \ \bm{CD}$. For comparison the mode without residual weight optimization is included}
\label{fig:1000YLoss}
\end{figure}

\begin{figure}[ht]
\centering
\includegraphics[scale=0.55]{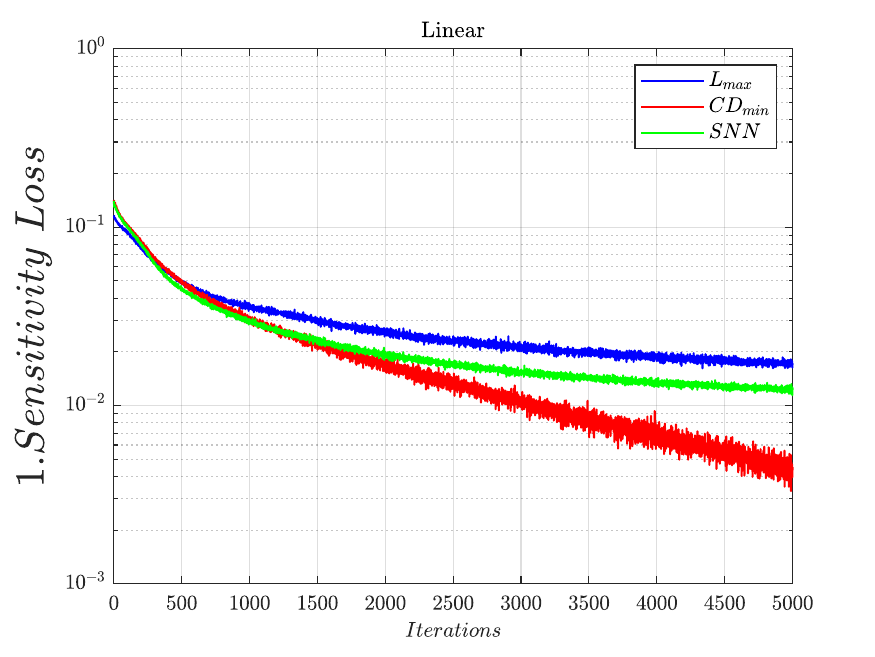} \
\includegraphics[scale=0.55]{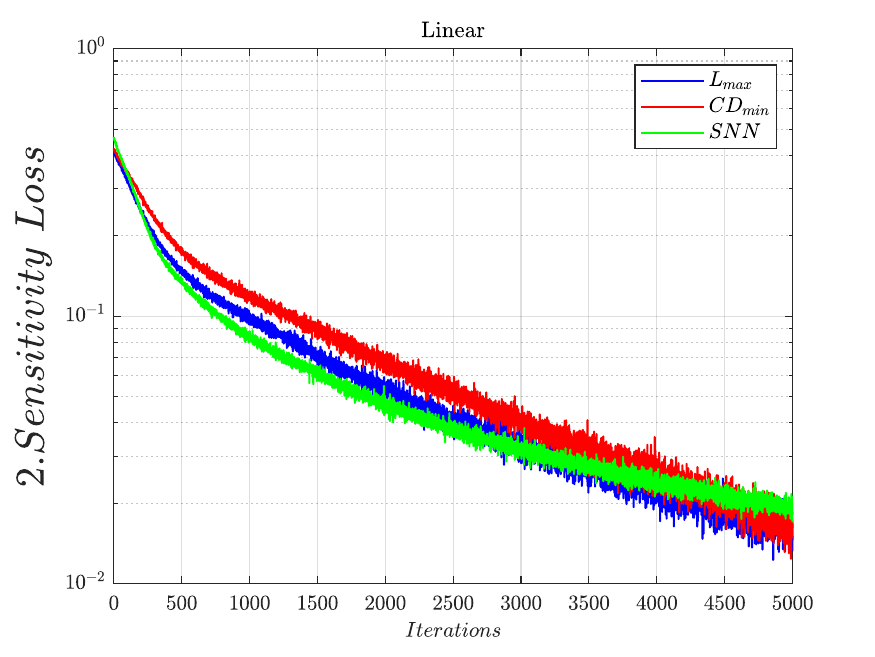}
\caption{Top: First sensitivity loss over the training iterations of $max \ \mathcal{L}$ and $min \ \bm{CD}$. For comparison the mode without residual weight optimization is included; Bottom: Second sensitivity loss over the training iterations of $max \ \mathcal{L}$ and $min \ \bm{CD}$. For comparison the mode without residual weight optimization is included}
\label{fig:1000dYLoss}
\end{figure}

\begin{figure}[ht]
\centering
\includegraphics[scale=0.55]{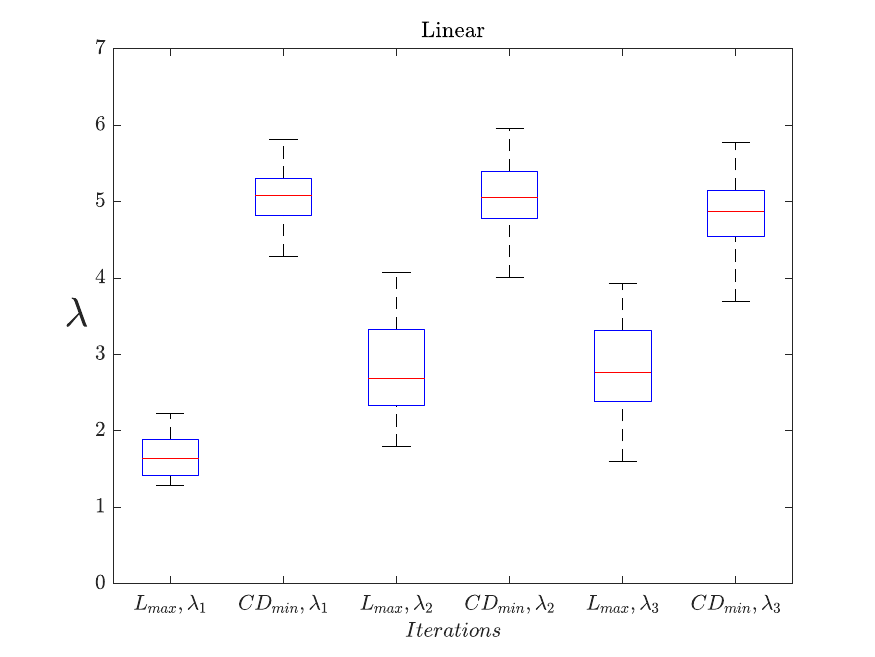}
\caption{Spread of the final residual weight values of all training runs for $max \ \mathcal{L}$ and $min \ \bm{CD}$}
\label{fig:1000rw}
\end{figure}

\begin{table}[ht]
\centering
\caption{Average final $l_2$ error for each mode. The averaged training time is included as well.}
\renewcommand{\arraystretch}{1.2}
\begin{tabular}{|c | c | c | c|} 
 \hline
 Mode & $L_2$ Error & Training Time [s]\\ [0.1ex] 
 \hline\hline
  1 & $0.111$ & $78$ \\ [0.1ex] 
 \hline
  2 & $0.076$ & $78$ \\ [0.1ex] 
 \hline
  3 & $0.115$ & $125$ \\ [0.1ex] 
 \hline
 4 & $0.071$ & $126$ \\ [0.1ex] 
 \hline
 5 & $0.085$ & $83$ \\ [0.1ex] 
 \hline
 6 & $0.057$ & $198$ \\ [0.1ex] 
 \hline
 7 & $0.256$ & $199$ \\ [0.1ex] 
 \hline
  8 & $0.073$ & $160$ \\ [0.1ex] 
 \hline
  9 & $0.093$ & $161$ \\ [0.1ex] 
 \hline
 10 & $0.086$ & $68$ \\ [0.1ex] 
 \hline
 11 & $0.094$ & $68$ \\ [0.1ex] 
 \hline
 12 & $0.078$ & $68$ \\ [0.1ex] 
 \hline
 13 & $0.085$ & $68$ \\ [0.1ex] 
 \hline
\end{tabular}
\label{table:500avgL2}
\end{table}
Following these results, modes 2 and 6 are chosen to be trained with 1000 epochs.
While mode 8 also shows promising results, the optimization method is similar to the better performing mode 6 and is therefore not continued for 1000 epochs.
In addition, the basic Sobolev model, mode 9, is trained with 1000 epochs for comparison.
This results in the loss and $l_2$ error curves over the iterations are visualised in figure \ref{fig:1000LossandL2}.
For further patterns, the individual losses of response and the two sensitivities are plotted in figures \ref{fig:1000YLoss}, \ref{fig:1000dYLoss}. 
It is to note that these individual losses do not include the residual weights, which is why the mode without residual weight optimization still has the lowest loss in figure \ref{fig:1000LossandL2}.
Lastly, the final residual weight values themselves are plotted for comparison in figure \ref{fig:1000rw}.

\subsection{Non-Linear Case}
\label{sec:Non-Linear Case}
\begin{figure}[ht]
\centering
\includegraphics[scale=0.55]{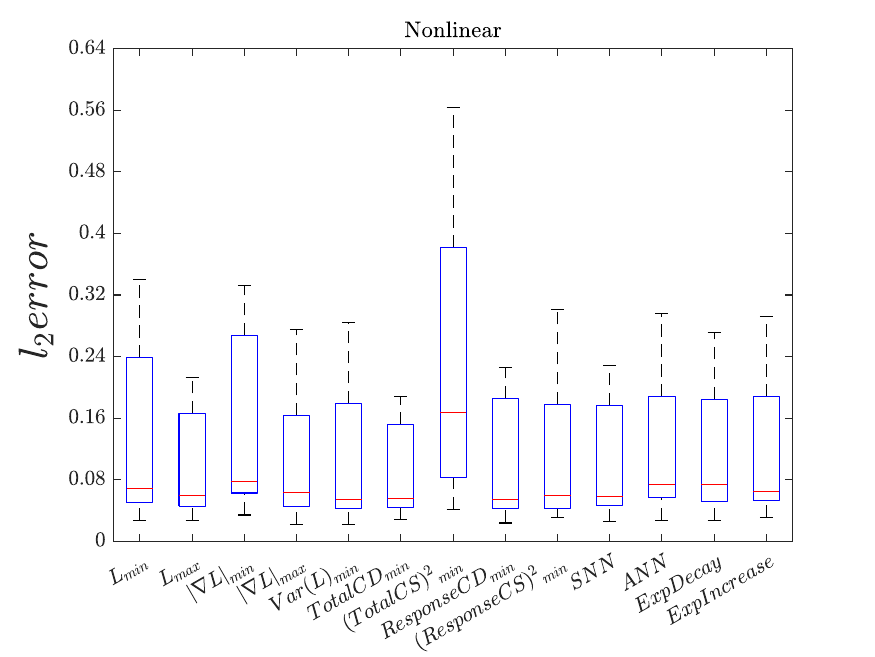}
\caption{$L_2$ error spread of all training runs for each mode}
\label{fig:nl500BestL2}
\end{figure}

\begin{figure}[ht]
\centering
\includegraphics[scale=0.5]{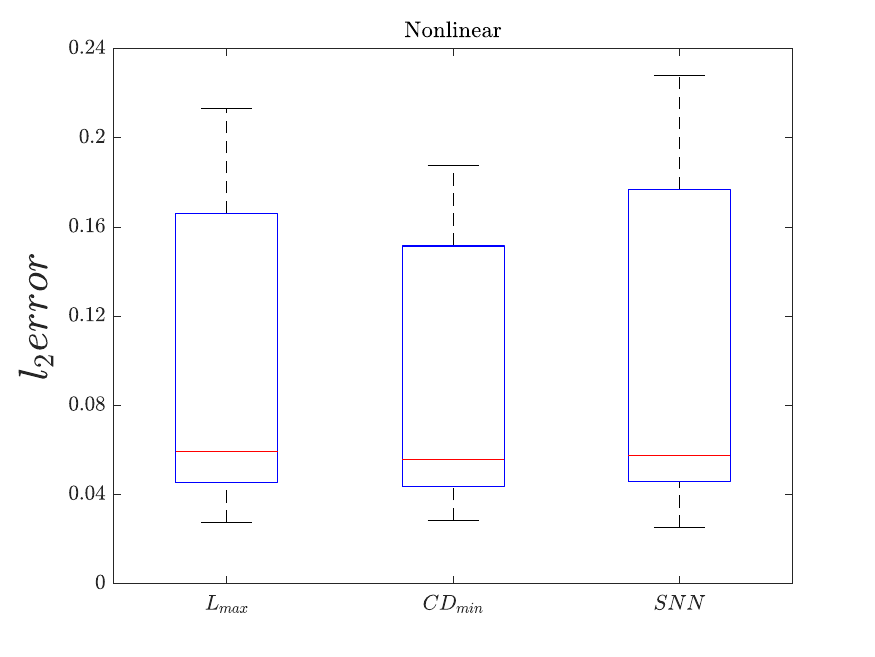}
\caption{$L_2$ error spread of $max \ \mathcal{L}$ and $min \ \bm{CD}$. For comparison the mode without residual weight optimization is included}
\label{fig:nl500L2}
\end{figure}

\begin{figure}[ht]
\centering
\includegraphics[scale=0.55]{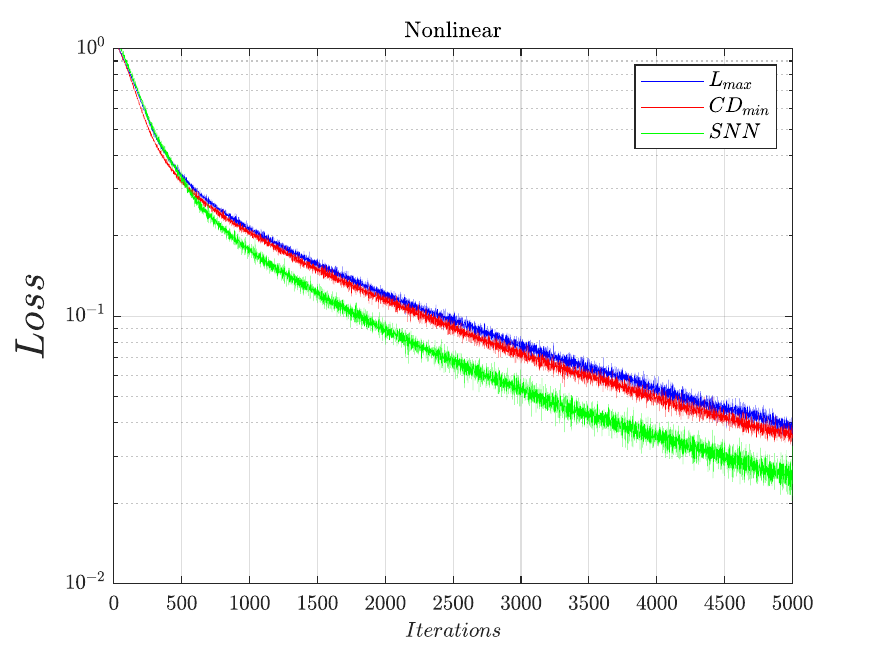} \
\includegraphics[scale=0.55]{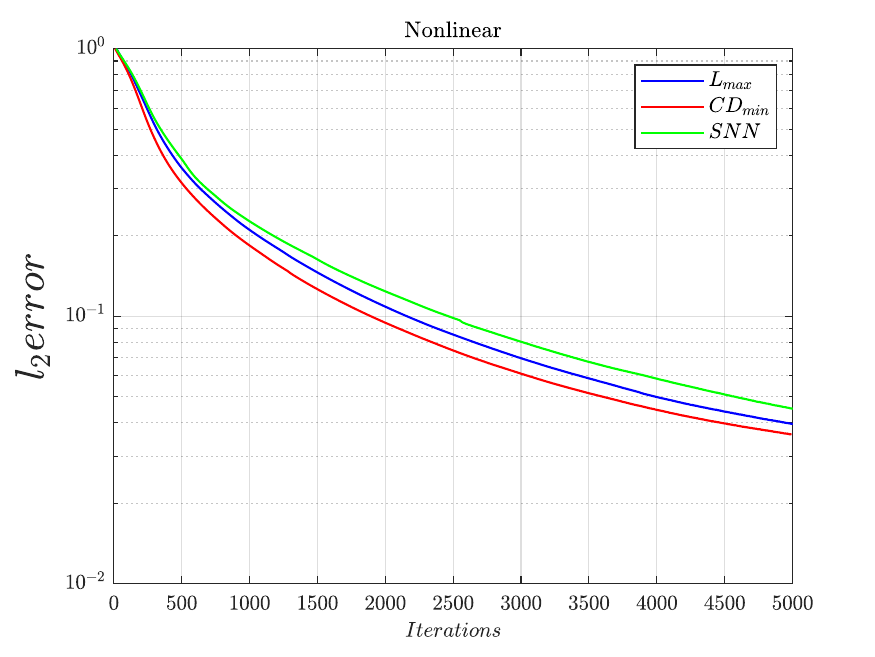}
\caption{Top: Loss over the training iterations of $max \ \mathcal{L}$ and $min \ \bm{CD}$. For comparison the mode without residual weight optimization is included; Bottom: $L_2$ error over the training iterations of $max \ \mathcal{L}$ and $min \ \bm{CD}$. For comparison the mode without residual weight optimization is included}
\label{fig:nl1000LossandL2}
\end{figure}

\begin{figure}[ht]
\centering
\includegraphics[scale=0.55]{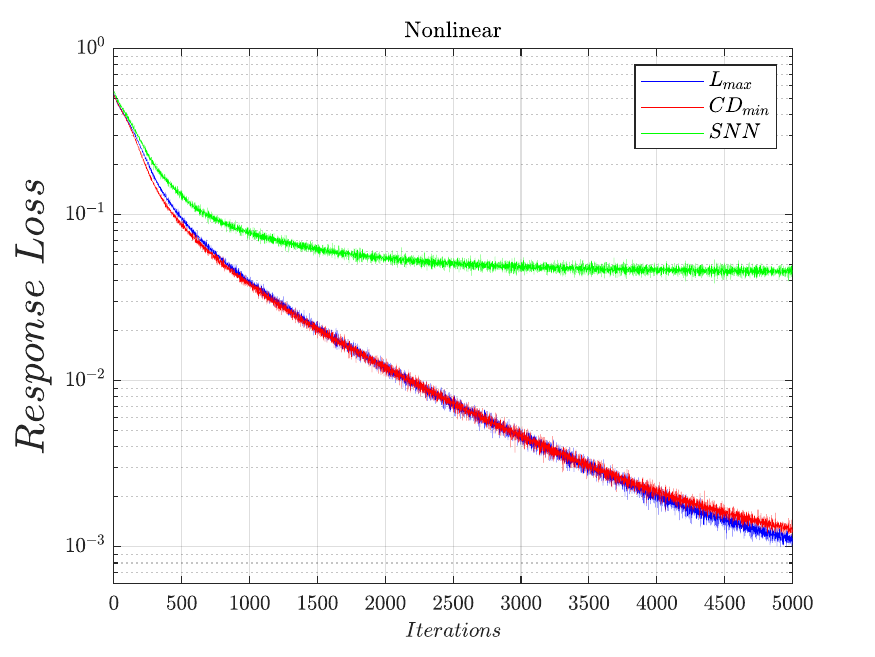}
\caption{Response loss over the training iterations of $max \ \mathcal{L}$ and $min \ \bm{CD}$. For comparison the mode without residual weight optimization is included}
\label{fig:nl1000YLoss}
\end{figure}

\begin{figure}[ht]
\centering
\includegraphics[scale=0.55]{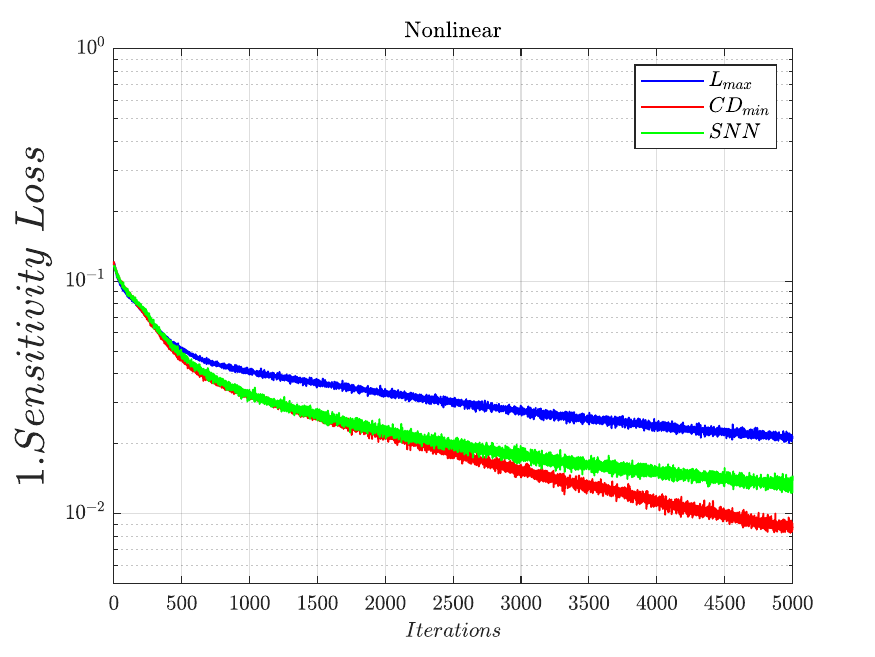} \
\includegraphics[scale=0.55]{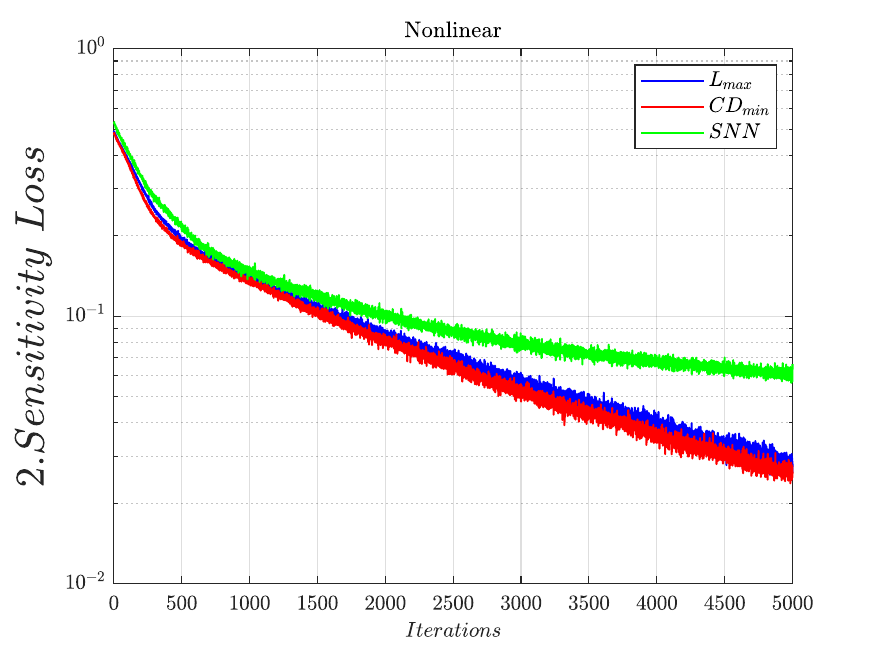}
\caption{Top: First sensitivity loss over the training iterations of $max \ \mathcal{L}$ and $min \ \bm{CD}$. For comparison the mode without residual weight optimization is included; Bottom: Second sensitivity loss over the training iterations of $max \ \mathcal{L}$ and $min \ \bm{CD}$. For comparison the mode without residual weight optimization is included}
\label{fig:nl1000dYLoss}
\end{figure}

\begin{figure}[ht]
\centering
\includegraphics[scale=0.55]{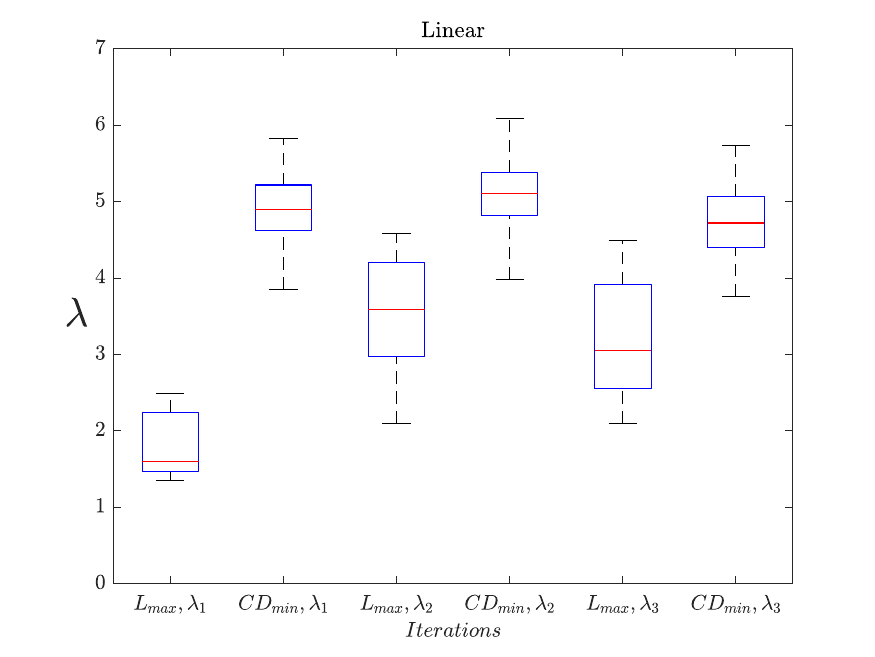}
\caption{Spread of the final residual weight values of all training runs for $max \ \mathcal{L}$ and $min \ \bm{CD}$}
\label{fig:nl1000rw}
\end{figure}

\begin{figure}[ht]
\centering
\includegraphics[scale=0.55]{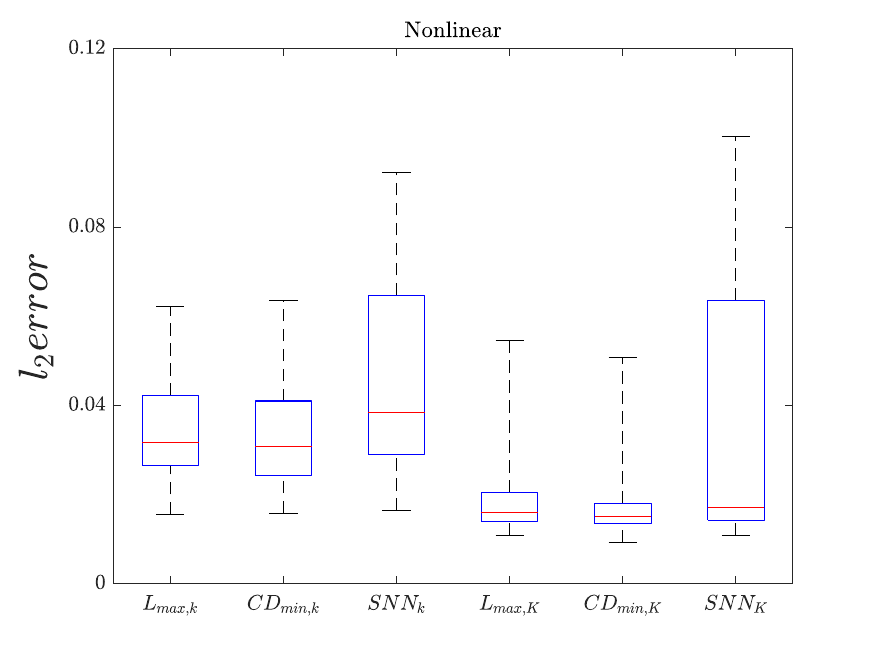}
\caption{Spread of the $l_2$ error values of all training runs for layer structures $5-3-3$ and $10-5-5-3$ for $max \ \mathcal{L}$ and $min \ \bm{CD}$}
\label{fig:nlBestL2layers}
\end{figure}

\begin{figure}[ht]
\centering
\includegraphics[scale=0.55]{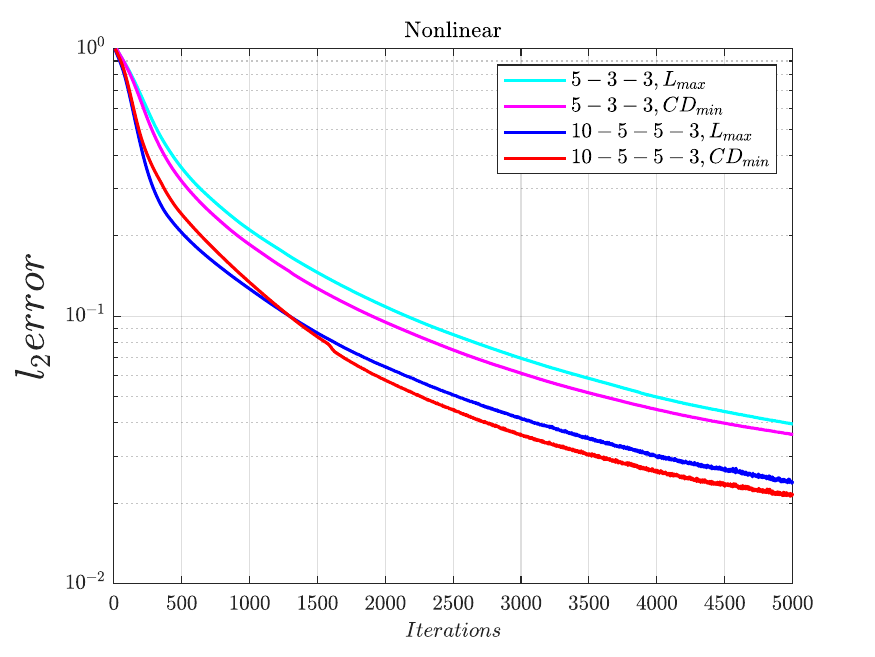}
\caption{Curve of the $l_2$ error over the iterations for layer structures $5-3-3$ and $10-5-5-3$ for $max \ \mathcal{L}$ and $min \ \bm{CD}$}
\label{fig:nlL2layers}
\end{figure}

For the non-linear elasticity case, training the models for 500 epochs results in the box plots in figure \ref{fig:nl500BestL2} for the $l_2$ error values.
For better analysis, the $l_2$ error box-plots of mode 2 and 6 are represented together with mode 9 for comparison, see figure \ref{fig:nl500L2}.
In addition the averaged $l_2$ error and training time results for each method are listed in table \ref{table:nl500avgL2}.
\begin{table}[ht]
\caption{Averaged final $l_2$ error values for each mode. The averaged training time is included as well.}
\centering
\renewcommand{\arraystretch}{1.2}
\begin{tabular}{|c | c | c | c|} 
 \hline
 Mode & $L_2$ Error & Training Time [s]\\ [0.1ex] 
 \hline\hline
  1 & $0.115$ & $78$ \\ [0.1ex] 
 \hline
  2 & $0.088$ & $78$ \\ [0.1ex] 
 \hline
  3 & $0.134$ & $124$ \\ [0.1ex] 
 \hline
 4 & $0.093$ & $123$ \\ [0.1ex] 
 \hline
 5 & $0.09$ & $80$ \\ [0.1ex] 
 \hline
 6 & $0.087$ & $191$ \\ [0.1ex] 
 \hline
 7 & $0.213$ & $197$ \\ [0.1ex] 
 \hline
  8 & $0.091$ & $144$ \\ [0.1ex] 
 \hline
  9 & $0.092$ & $144$ \\ [0.1ex] 
 \hline
 10 & $0.089$ & $67$ \\ [0.1ex] 
 \hline
 11 & $0.112$ & $68$ \\ [0.1ex] 
 \hline
 12 & $0.103$ & $67$ \\ [0.1ex] 
 \hline
 13 & $0.105$ & $67$ \\ [0.1ex] 
 \hline
\end{tabular}
\label{table:nl500avgL2}
\end{table}
Following these results, modes 2 and 6 are chosen to be trained with 1000 epochs.
In addition, the basic Sobolev model, mode 9, is trained with 1000 epochs for comparison.
This results in loss and $l_2$ error curves over the iterations in figure \ref{fig:nl1000LossandL2}.
For further patterns, the individual losses of response and the two sensitivities are plotted in figures \ref{fig:nl1000YLoss}, \ref{fig:nl1000dYLoss}.
It is to note that these individual losses do not include the residual weights, which is why the mode without residual weight optimization has the lowest loss in figure \ref{fig:nl1000LossandL2}.
Lastly, the final residual weight values themselves are plotted for comparison in figure \ref{fig:nl1000rw}.
For closing, results for these same models, trained on 2000 epochs, are produced to observe the effects of the epoch length on the training convergence.
\begin{table}[ht]
\caption{Average and lowest $l_2$ error for $max \ \mathcal{L}$, $min \ \bm{CD}$ and $SNN$ for different number of epochs}
\centering
\renewcommand{\arraystretch}{1.2}
\begin{tabular}{|c | c | c | c|} 
 \hline
 Mode & Epochs & Avg. $L_2$ Error [s] & Smallest $L_2$ Error\\ [0.5ex] 
 \hline\hline
  $max \ \mathcal{L}$ & $500$ & $0.088$ & $0.027$\\ [0.5ex] 
 \hline
  $min \ \bm{CD}$ & $500$ & $0.087$ & $0.028$\\ [0.5ex] 
 \hline
  $SNN$ & $500$ & $0.089$ & $0.025$\\ [0.5ex] 
 \hline\hline
 $max \ \mathcal{L}$ & $1000$ & $0.039$ & $0.015$\\ [0.5ex] 
 \hline
 $min \ \bm{CD}$  & $1000$ & $0.036$ & $0.016$\\ [0.5ex] 
 \hline
 $SNN$ & $1000$ & $0.045$ & $0.016$\\ [0.5ex] 
 \hline\hline
 $max \ \mathcal{L}$ & $2000$ & $0.026$ & $0.014$\\ [0.5ex] 
 \hline
 $min \ \bm{CD}$  & $2000$ & $0.026$ & $0.014$\\ [0.5ex] 
 \hline
  $SNN$ & $2000$ & $0.025$ & $0.013$\\ [0.5ex] 
 \hline
\end{tabular}
\label{table:nl1000v2000avgandMinL2}
\end{table}
The table \ref{table:nl1000v2000avgandMinL2} shows the values for 500, 1000 and 2000 epochs.
Furthermore, the effects of the model layer size is observed with additional results for a layer structure of 10-5-5-3 and compared to the original 5-3-3 model in figures \ref{fig:nlBestL2layers} and \ref{fig:nlL2layers}.
Lastly, the table \ref{table:nlavgandMinBestL2layers} illustrates the difference of the average and minimum $l_2$ error for both layer structures.
\begin{table}[ht]
\centering
\caption{Average and lowest $l_2$ error for $max \ \mathcal{L}$ and $min \ \bm{CD}$ for the two different layer sizes}
\renewcommand{\arraystretch}{1.2}
\begin{tabular}{|c | c | c | c|} 
 \hline
 Mode & Layers & Avg. $L_2$ Error & Lowest $L_2$ Error\\ [0.5ex] 
 \hline\hline
  $max \ \mathcal{L}$ & $5-3-3$ & $0.035$ & $0.016$\\ [0.5ex] 
 \hline
  $min \ \bm{CD}$ & $5-3-3$ & $0.034$ & $0.016$\\ [0.5ex]
 \hline\hline
 $max \ \mathcal{L}$ & $10-5-5-3$ & $0.022$ & $0.011$\\ [0.5ex] 
 \hline
 $min \ \bm{CD}$ & $10-5-5-3$ & $0.020$ & $0.009$\\ [0.5ex] 
 \hline
\end{tabular}
\label{table:nlavgandMinBestL2layers}
\end{table}

\section{Analysis and Discussion}
\label{sec:AnalysisandDiscussion}
While the analysis will be split into linear and non-linear case, the non-linear case will have a slightly expanded analysis.
This is because when observing the first results in the non-linear case, the differences between the best performing modes is less pronounced, as will be discussed later.
This is not the case for the linear results.
The non-linear case discussion therefore also cover the 2000 epochs training results.
In addition the effect of the expansion of layers is only analysed for the non-linear case.

\subsection{Linear Case}
\label{sec:LinearCase}
The linear case results show decisively, which modes performs best. Figure \ref{fig:500BestL2} shows the spread of the multiple runs of the different applied modes of training.
It can be observed that mode 6 performs best.
Following this, modes 2 and 4 perform second best.
The other modes seem to perform somewhat equal, with modes 1, 3 and 7 performing the worst.
In figure \ref{fig:500L2}, the difference in performance of mode 2 and 6, as well as the base performance of the Sobolev trained model, mode 10, becomes clearer.
The gradient alignment applied in mode 6 shifts the spread of the $l_2$ error slightly further down but also increases the precision of the spread of values in comparison to mode 2 and 10.
This can also be observed in the average and median being lower than the other two modes, as well as the lowest obtained value being better.
In table \ref{table:500avgL2} the difference of performance is visualised with the average scalar $l_2$ error values.
First, mode 6 is clearly the best performing mode, followed by mode 4 and 2.
However when considering the average training time for the modes, mode 6 is second to last.
In comparison, mode 2 takes less than half the time and mode 4 only takes ca. $60\%$ to train.
While the loss itself is no longer a good metric for comparison of modes because of the Sobolev training, the modes 2 and 6 have a bigger loss than the basic Sobolev training, see figure \ref{fig:1000LossandL2}.
To actually judge the performance difference, figure \ref{fig:1000LossandL2} provides the $l_2$ error, where both modes outperform the basic Sobolev training mode, meaning it takes less time to train a model for a certain accuracy.
It can also be observed that modes 2 and 6 perform relatively equal until ca. iteration $2000$, where mode 6 begins to outperform mode 2 by a slight margin.
Both of these figures are the results of 1000 epoch training.
Compared to the 500 epoch training, there is a performance difference in magnitude.
However, when it comes to modes 2 and 6, which were performing the best, the difference in performance is unchanged.
Interestingly, when observing the results in figures \ref{fig:1000YLoss}, \ref{fig:1000dYLoss}, mode 6 reduces the losses of all individual loss addends the best.
Mode 2 seems to have difficulties approximating the first sensitivity and the basic Sobolev training mode only keeps up the performance for the second sensitivity.
Still, for the total loss, the residual weights are included in the calculation, which leads to the basic Sobolev training mode having the lowest loss of all three, as mentioned for figure \ref{fig:1000LossandL2}.
When analysing the values the residual weights end up with, it can be seen in figure \ref{fig:1000rw} that mode 2 generally converges the residual weights to lower ranges than mode 6.
Additionally, while all three residual weights of mode 6 range in similar regions, in the case of mode 2, the residual weight of the response loss is generally in a lower range than the residual weights of the sensitivity losses.
In both cases, the residual weight of the response loss seems to have the highest precision.

\subsection{Non-Linear Case}
\label{sec:Non-LinearCase}
The non-linear case results show how the performance between modes is not as clear as the linear case. Figure \ref{fig:nl500BestL2} shows the spread of the multiple runs of the different applied modes of training.
It can be observed that for the non-linear case mode 6 performs best as well.
Following this, modes 2 and 4 perform second best.
The other modes seem to perform somewhat equal, with modes 1, 3 and 7 performing the worst.
However the difference in performance is not quite as big as it was in the linear case.
Modes 2, 4 and 6 only outperform slightly.
In figure \ref{fig:nl500L2}, the difference in performance of mode 2 and 6, as well as the base performance of the Sobolev trained model, mode 10, becomes clearer.
The gradient alignment applied in mode 6 increases the precision of the spread of values in comparison to mode 2 and 10.
However there is no strong performance improvement for the average and median, as well as the lowest obtained $l_2$ error.
In table \ref{table:nl500avgL2} the difference of performance is visualised with the average scalar $l_2$ error values.
First, mode 6 is slightly outperforming the other modes, but not substantially.
Surprisingly, mode 10, which is the basic Sobolev training mode, performs equally well, meaning unadjusted residual weights seem to perform just as well as the various applied methods.
This is quite different compared to the linear results.
The average training time of the modes behaves similar to the linear case.
Mode 6 is the second to last longest training mode, with mode 2 and 4 showing similar ratios to it as in the linear case.
In addition, mode 10 performing equally well, takes the lowest amount of training time.
This very different result compared to the linear case is the reason for the additional analysis by expansion of epochs and layer structure.
While the loss itself is no longer a good metric for comparison of modes because of the Sobolev training, the modes 2 and 6 have a bigger loss than the basic Sobolev training, see figure \ref{fig:nl1000LossandL2}.
To actually judge the performance difference, figure \ref{fig:nl1000LossandL2} provides the $l_2$ error, where both modes outperform the basic Sobolev training mode, meaning again it takes less time to train a model for a certain accuracy with these modes.
It can also be seen that with 1000 epochs the modes 2 and 6 outperform the basic Sobolev training mode.
In addition, mode 6 outperforms mode 2.
Compared to the 500 epoch training, there is again a performance difference, which will be discussed later in this section, in comparison to the 2000 epoch results.
Interestingly, when observing the results in \ref{fig:nl1000YLoss}, \ref{fig:nl1000dYLoss}, mode 6 reduces the losses of all individual loss addends the best again, just like the linear case.
Mode 2 seems to again have difficulties approximating the first sensitivity. 
Furthermore, the basic Sobolev training mode now seems to have difficulties keeping up performance of these three metrics.
Still again, for the total loss, the residual weights are included in the calculation, which leads to the basic Sobolev training mode having the lowest loss of all three, as mentioned in figure \ref{fig:nl1000LossandL2}.
When analysing the values the residual weights end up with for the non-linaer acse, it can be seen in figure \ref{fig:nl1000rw} that mode 2 again generally converges the residual weights to lower ranges than mode 6.
In general, the same behavior mentioned for the linear case is also found here for the non-linear case.
The residual weights also stay in the same magnitude as the linear case.
However the residual weights for the sensitivities for mode 2 seems to have a slightly bigger spread.
Additionally, while mode 2 has the highest precision for the first residual weight, mode 6 seems to have similar behavior of precision for all three residual weights.
It can be observed in table \ref{table:nl1000v2000avgandMinL2} that the final accuracy improves with epoch size.
But the performance difference between $max \ \mathcal{L}$ and $min \ \bm{CD}$ does not seem to change with the epoch size.
However, basic Sobolev training, substantially increases its performance with the epoch size relative to the other two modes.
Quite more interesting are the results for the expanded model layer structure.
When considering figure \ref{fig:nlBestL2layers}, both mode 2 and 6 perform better.
It can be noticed that the average and median $l_2$ error has improved better for the second layer structure "10-5-5-3" in both cases, however the difference between mode 2 and 6 seems to have expanded.
Furthermore, the lowest obtained $l_2$ error has also improved for mode 6 in comparison to mode 2.
The general superior performance of mode 2 compared to mode 6 over the training process can also be observed in figure \ref{fig:nlL2layers}.
Generally, mode 6 produces better accuracy than mode 2 over the training iterations.
Especially of note is table \ref{table:nlavgandMinBestL2layers}, where the performance difference between mode 2 and 6 can be seen to be expanding for a bigger layer structure.
This shows, that the small model dimensions seem to bottleneck the performance of the gradient alignment method.
Therefore, if the model were to be sufficiently sized, more than likely, the same behavior of results as the linear case should be obtained for the non-linear case.

\section{Conclusion}
\label{sec:Conclusion}
In this paper we applied the generated data from a FEM simulating the behavior of stress triaxiality response, as well as its sensitivitiy w.r.t. chosen geometric paramters, of a linear and nonlinear material hook-shaped object under a defined set of force onto a neural network, making use of sensitivities. 
From here on various methods of finding the optimal set of residual weights applied to the individual losses of the gradient-enhanced neural network were observed. 
During comparison it was concluded that the method of minimising the cosine distance to align the gradient vectors of each individual loss with the total loss showed best convergence performance. 
Furthermore this gradient alignment method showed further improvement with expansion of layer width and depth. 
Specifically, the gradient alignment reduced the spread with an increased accuracy of the neural network training compared to Sobolev training. 
This was the case for both the linear and nonlinear material data.
Still, concerning the training time of the neural network, the gradient alignment method performed the worst.
While the idea behind gradient alignment seems to work, there is great room for improvement in regards to optimization of the training algorithms computational efficiency.
As a matter of fact, a more optimized and faster training variation has been tried and tested with similar accurate results.
Another point of interest is the expansion of the layer structure, which could act as a bottleneck for the performance of the gradient alignment method.
Finally, with respect to the applied method of minimising the cosine distance to optimise the residual weights $\lambda$, it is understood that the residual weight adjusts the magnitude of the gradient vectors to affect the angle between the individual losses and the total loss.
From this, introducing residual weights $\bm{\lambda}$ for each element of the gradient vectors would allow the algorithm to adjust not only magnitude but also the direction of the gradient vector.
Thereby, the algorithm would ideally discard elements of gradient vectors which oppose each other and amplify elements that go hand in hand.
However following that thought, the algorithm method would need to be reconsidered, as potentially minimising the cosine distance would have undesired effects of reducing convergence performance.
Additionally, depending on the size of the neural network, having residual weights for each element results in more computational effort needed and should be considered carefully.
Lastly, in the early results, the method aligning gradient vectors to the response only showed good results as well, which could be considered for mixing with other methods.

\newpage

\interlinepenalty=10000
\bibliographystyle{elsarticle-num}
\bibliography{literature}

\end{document}